\documentclass[journal]{IEEEtran}
\usepackage{amssymb}

\usepackage{multicol}
\usepackage{pifont,calc}
\usepackage{cite}

\usepackage{graphicx}
\ifCLASSINFOpdf
   \usepackage[pdftex]{graphicx}
\else
 \fi
\usepackage[cmex10]{amsmath}
\usepackage{array}

\usepackage{cases} 
\usepackage{bm} 
\usepackage{subfigure}
\usepackage{mathrsfs}
\usepackage{url}
\usepackage{algorithm}
\usepackage{algpseudocode}

\usepackage{arydshln}
\usepackage{multirow}
\usepackage{multicol}
\makeatletter

\newcommand{\Rmnum}[1]{\expandafter\@slowromancap\romannumeral #1@}
\makeatother

\ifCLASSINFOpdf
\else
\fi
\begin{document}
%
\title{Low-Memory Implementations of Ridge Solutions  for Broad Learning System with Incremental Learning}
%
%
%

\author{Hufei~Zhu
\thanks{H. Zhu is with Faculty of Intelligent Manufacturing, Wuyi University, Jiangmen 529020, China (e-mail:
zhuhufei@wyu.edu.cn).}}

%
%

\markboth{Journal of \LaTeX\ Class Files,~Vol.~14, No.~8, August~2015}%
{Shell \MakeLowercase{\textit{et al.}}: Bare Demo of IEEEtran.cls for IEEE Journals}
%

%
%
%
%
%



\maketitle

\begin{abstract}
The existing low-memory BLS implementation   proposed recently
 avoids the need for storing and
inverting large matrices,
to achieve efficient usage of memories.
 However, the existing low-memory BLS implementation sacrifices the  testing accuracy as a price
for efficient usage of memories, since
 it  can no longer obtain the generalized inverse or ridge solution
for the output weights during incremental learning, and it cannot work
under  the very small ridge parameter  (i.e., $\lambda = 10^{-8}$) that is utilized in the original BLS.
 Accordingly, it is required to develop the low-memory BLS implementations,
   which can work  under very small
 ridge parameters and compute the generalized inverse or ridge solution
for the output weights
in the process of incremental learning.

In this paper, firstly we propose the low-memory  implementations
for the  recently proposed recursive and square-root BLS algorithms on added inputs and
 the  recently proposed  square-root BLS algorithm on added nodes, by  simply processing a batch of inputs or nodes in each
 recursion.  Since the recursive BLS implementation includes
 the recursive updates of
       the inverse matrix that
may introduce numerical instabilities
   after a large number of iterations,
   and needs the extra computational load to decompose the
 inverse matrix  into the Cholesky factor  when cooperating with the proposed low-memory implementation of
 the square-root BLS algorithm on added nodes,
 we only improve the  low-memory implementations of the square-root BLS algorithms on added inputs and nodes,
 to propose the full low-memory implementation of the square-root BLS algorithm.

  All the proposed low-memory BLS  implementations  compute the  ridge solution
for the output weights
in the process of incremental learning, and most of them can work  under very small
 ridge parameters.
  When the ridge parameter is not too small, the proposed low-memory  implementations for
the recursive and square-root BLS algorithms on added inputs and the part for added inputs
of the proposed full low-memory BLS implementation
 usually achieve  better testing accuracies than the existing low-memory  BLS implementation on added inputs.
 More importantly,
 when the ridge parameter
  is very small (i.e., $\lambda = 10^{-8}$) as in the original BLS,
  the existing low-memory  BLS implementation on added inputs  cannot work in any update (of the incremental learning),
 the proposed low-memory  implementation for the recursive BLS algorithm on added inputs
  cannot work in the last update,
  while  the proposed low-memory  implementation for
the square-root BLS algorithm on added inputs and the proposed full low-memory implementation of the square-root BLS algorithm (on added inputs and nodes)
   can work in all updates.

    With respect to the existing low-memory  BLS implementation on added inputs,
  the proposed low-memory  implementation for
the recursive BLS algorithm on added inputs and  the part for added inputs
of the proposed full low-memory implementation of the square-root BLS algorithm
    require nearly
 the same training time, while the proposed low-memory  implementation for
the square-root BLS algorithm on added inputs requires much more training time. On the other hand,
   the part for added nodes of  the proposed full low-memory implementation of the square-root BLS algorithm speeds up
 the  proposed low-memory  implementation
for the square-root BLS algorithm on new added nodes
   by a factor in the range from $3.77$ to $23.73$.

The  proposed  full low-memory implementation of the square-root BLS algorithm on added inputs and nodes
can work when the ridge parameter
  is very small (i.e., $\lambda = 10^{-8}$) as in the original BLS.
The part for added inputs of  the  proposed  full low-memory implementation of the square-root BLS algorithm
  takes nearly
 the same training time to
   achieve
 better testing accuracies
   with respect to  the existing low-memory  BLS implementation on added inputs,
    is numerically more stable  than  the proposed low-memory  implementation for
the recursive BLS algorithm on added inputs,
 and is much faster than  the proposed low-memory  implementation for
the square-root BLS algorithm on added inputs.
On the other hand, the part for added nodes of  the  proposed  full low-memory implementation of the square-root BLS algorithm  is obviously faster than
the  proposed low-memory  implementation
for the square-root BLS algorithm on added nodes.
Accordingly,
it can be concluded that
  the  proposed  full low-memory implementation of the square-root BLS algorithm on added inputs and nodes
  is
  a good low-memory  implementation of the original BLS on added
 nodes and inputs.
\end{abstract}


%
\IEEEpeerreviewmaketitle

\section{Introduction}

For the  classification and
regression problems,
Single layer feedforward neural networks (SLFN)  have been
widely applied~\cite{BL_Ref_18,BL_Ref_19,BL_Ref_20}.
SLFN can be trained by
traditional  gradient-descent algorithms~\cite{BL_Ref_22,BL_Ref_23},
which are usually  time-consuming and require huge
processing power. Accordingly, the random vector functional-link neural network (RVFLNN)
has been utilized to
 overcome the
need for gradient-descent training~\cite{BL_Ref_19}.

To model  time-variety data with moderate size,
a dynamic step-wise updating
algorithm was proposed in \cite{27_ref_BL_trans_paper} for the RVFLNN model,
which includes the concept of ``incremental learning,"
i.e.,  remodelling  the network in an incremental way without a complete retraining process.
When a new input is added or a new node is inserted, the step-wise updating algorithm in \cite{27_ref_BL_trans_paper}
remodels the network by only computing the pseudoinverse of that added input or node.
Recently to deal with time-variety big data with
high dimension,
 Broad Learning
System (BLS) was proposed in \cite{BL_trans_paper},
which improves the step-wise updating algorithm in \cite{27_ref_BL_trans_paper}.
In BLS,  the input data is transformed into the feature nodes
  to reduce the data dimensions, while
  the output weights are computed by the generalized inverse with the ridge regression approximation to
  achieve a better generalization performance, which
 assumes the ridge parameter $\lambda \to 0$ in the ridge inverse~\cite{best_ridge_inv_paper213}.

 To improve the original BLS on new added inputs in \cite{BL_trans_paper},
the inverse of a sum of matrices~\cite{InverseSumofMatrix8312} was utilized
in \cite{my_brief1_on_BL},
to accelerate a step in the generalized inverse of a  row-partitioned matrix. Moreover,
 two improved BLS algorithms
were proposed in \cite{mybrief2onBLInputs2019}
to further accelerate the BLS algorithm on new added inputs,
which compute the ridge solution~\cite{best_ridge_inv_paper213} for the output weights
   to achieve a better generalization performance,
   instead of the generalized inverse solution with the ridge regression approximation
   utilized in the original BLS~\cite{BL_trans_paper}.
Accordingly  in \cite{mybrief2onBLInputs2019}, it is no longer required to assume
the ridge parameter $\lambda \to 0$ in the ridge inverse, and $\lambda$ can be any positive real number.
To reduce the computational complexity,
    the recursive BLS algorithm and    the square-root BLS  algorithm
    proposed in \cite{mybrief2onBLInputs2019}
    compute the output weights from the inverse and the inverse Cholesky factor of
  the Hermitian matrix in the ridge inverse, respectively,
  which are usually smaller than the ridge inverse.
  On the other hand,   the square-root BLS  algorithms on  added nodes have been proposed
  in \cite{CholBLSdec2020, mybrief1onBLNodes2019} to improve the original BLS on added nodes in \cite{BL_trans_paper}.
  The square-root BLS  algorithm proposed in \cite{CholBLSdec2020} still computes  the generalized inverse solution for the output weights,
  while in \cite{mybrief1onBLNodes2019},  the square-root BLS  algorithm has been proposed to
   compute the ridge solution for the output weights from the inverse Cholesky factor of
  the Hermitian matrix in the ridge inverse.

 Recently, a low-memory implementation of BLS was proposed
 in \cite{BLSLowMemTNNLS2020dec}
for big-data scenarios,
 which
 avoids the need for storing and
inverting large matrices, and finds a good tradeoff between  efficient usage of memories and
required iterations. However, the low-memory incremental learning algorithm  
proposed in \cite{BLSLowMemTNNLS2020dec} can no longer obtain the generalized inverse or ridge solution
for the output weights, and then it sacrifices the  testing accuracy as a price
for efficient use of memories,
with respect to the generalized inverse solution (with
the ridge regression approximation) in \cite{BL_trans_paper} and the ridge solution in \cite{mybrief2onBLInputs2019,mybrief1onBLNodes2019}.
On the other hand, the low-memory implementation in \cite{BLSLowMemTNNLS2020dec}
has not provided the simulations
under  very small
 ridge parameters as the original BLS~\footnote{The simulations in \cite{BLSLowMemTNNLS2020dec}
 set the equivalent ridge parameter (i.e., the regularization factor)
 $\lambda = 1/128 \approx 0.0078$, while
  the original BLS~\cite{BL_trans_paper},
 which  is based on  the generalized inverse with the ridge regression approximation,
 assumes the ridge parameter $\lambda \to 0$
 and usually
needs to set
  a very small
 ridge parameter (e.g., $\lambda = 10^{-8}$ in \cite{BL_trans_paper}).} and the efficient BLS algorithms proposed in \cite{mybrief2onBLInputs2019,mybrief1onBLNodes2019}.
 Accordingly, it is still required to develop the low-memory BLS implementations,
   which can work  under very small
 ridge parameters and 
  compute the generalized inverse or ridge solution
for the output weights
in the process of incremental learning.



In this paper, firstly we propose the low-memory  implementations
for both
 BLS algorithms on added inputs proposed in \cite{mybrief2onBLInputs2019} and
 the  BLS algorithm on added nodes proposed in
 \cite{mybrief1onBLNodes2019}, by  simply processing a batch of inputs or nodes in each recursion.
 Then we propose the full low-memory implementation of the square-root BLS algorithm for added inputs and nodes,
  which is better than the above-mentioned low-memory BLS  implementations.
  All the proposed low-memory BLS  implementations  compute the  ridge solution
for the output weights
in the process of incremental learning, and most of them can work  under very small
 ridge parameters.


  This paper is organized as follows. Section \Rmnum{2} introduces the existing incremental BLS algorithms on added inputs and nodes.
  In Section \Rmnum{3}, we propose  the low-memory implementations of the BLS algorithms proposed in \cite{mybrief2onBLInputs2019,mybrief1onBLNodes2019}.
Then in Section \Rmnum{4}, we propose the full low-memory implementation of the square-root BLS algorithm on added inputs and nodes that is better than
the  implementations proposed  in Section \Rmnum{3}.
The presented low-memory BLS implementations are compared
by numerical experiments  in Section \Rmnum{5},
and
conclusions are given in Section \Rmnum{6} finally.

%
%

 \section{Existing Incremental BLS on Added Inputs  and Nodes}

%
%
%
%
%
%
%
%
%
%
%
%
%
%

In this section, we will introduce the model of BLS,
the generalized inverse solution and the ridge solution for BLS,
and  the incremental BLS algorithms on added inputs and nodes   proposed in  \cite{mybrief2onBLInputs2019} and \cite{mybrief1onBLNodes2019} ,
respectively.

\subsection{Broad Learning Model}




 In the BLS,  the original input data
 $\mathbf{X}\in {\Re ^{l \times q}}$ with $l$ training samples
 is projected by
\begin{equation}\label{Z2PhyXWb985498}
{{\mathbf{Z}}_{i}}=\phi (\mathbf{X}{{\mathbf{W}}_{{{e}_{i}}}}+{{\mathbf{\beta }}_{{{e}_{i}}}}),
\end{equation}
to become the $i$-th group of mapped features ${{\mathbf{Z}}_{i}}$,
where  the  weights ${{\mathbf{W}}_{{{e}_{i}}}}$ and the biases ${{\mathbf{\beta }}_{{{e}_{i}}}}$ are
  randomly generated and then fine-tuned by applying the linear inverse problem~\cite{BL_trans_paper}.
All the  $n$ groups of mapped features can be concatenated into
\begin{equation}\label{Zi2z1zi988689}{{\mathbf{Z}}^{n}}\equiv \left[ \begin{matrix}
   {{\mathbf{Z}}_{1}} & \cdots  & {{\mathbf{Z}}_{n}}  \\
\end{matrix} \right],
\end{equation}
 which are then enhanced by
\begin{equation}\label{HjipsenZjWbelta09885}{{\mathbf{H}}_{j}}=\xi ({{\mathbf{Z}}^{n}}{{\mathbf{W}}_{{{h}_{j}}}}+{{\mathbf{\beta }}_{{{h}_{j}}}}),
\end{equation}
 to become the
$j$-th group of enhancement nodes ${{\mathbf{H}}_{j}}$,
where
    ${{\mathbf{W}}_{{{h}_{j}}}}$ and ${{\mathbf{\beta }}_{{{h}_{j}}}}$   are randomly generated.
   All the $m$ groups of enhancement nodes can be concatenated into
\begin{equation}\label{Hj2H1Hj9859348}{{\mathbf{H}}^{m}}\equiv \left[ {{\mathbf{H}}_{1}},\cdots, {{\mathbf{H}}_{m}} \right].
\end{equation}
The above described procedure to initialize BLS
is summarized in \textbf{Algorithm 1} by the function
  \begin{equation}\label{}
 ({{\mathbf{Z}}^{n}}, {{\mathbf{H}}^{m}}, {{\mathbf{W}}_{{{e}}}^n }, {{\mathbf{\beta }}_{{{e}}}^n} , {{\mathbf{W}}_{{{h}}}^m }, {{\mathbf{\beta }}_{{{h}}}^m})=
\emph{Initialize} (\mathbf{X}).
 \end{equation}


\begin{algorithm}
\caption{The Procedure to Initialize BLS}\label{euclid}
\begin{algorithmic}[0]
\Function{${Initialize}$}{$\mathbf{X}$}
\State Compute ${{\mathbf{Z}}_{i}}=\phi (\mathbf{X} {{\mathbf{W}}_{{{e}_{i}}}}+{{\mathbf{\beta }}_{{{e}_{i}}}})$
with fine-tuned random ${{\mathbf{W}}_{{{e}_{i}}}}$ and ${{\mathbf{\beta }}_{{{e}_{i}}}}$ for $i=1:n$;
\State      Set   ${{\mathbf{Z}}^{n}}\equiv \left[ \begin{matrix}
   {{\mathbf{Z}}_{1}} & \cdots  & {{\mathbf{Z}}_{n}}  \\
\end{matrix} \right]$, $ {{\mathbf{W}}_{{{e}}}^n } \equiv[{{\mathbf{W}}_{{{e}_{1}}}} \cdots {{\mathbf{W}}_{{{e}_{n}}}}]$, ${{\mathbf{\beta }}_{{{e}}}^n} \equiv[{{\mathbf{\beta }}_{{{e}_{1}}}} \cdots {{\mathbf{\beta }}_{{{e}_{n}}}}]$;
\State Compute  ${{\mathbf{H}}_{j}}=\xi ({{\mathbf{Z}}^{n}}{{\mathbf{W}}_{{{h}_{j}}}}+{{\mathbf{\beta }}_{{{h}_{j}}}})$ with random ${{\mathbf{W}}_{{{h}_{j}}}}$ and ${{\mathbf{\beta }}_{{{h}_{j}}}}$ for $j=1:m$;
\State   Set ${{\mathbf{H}}^{m}}\equiv \left[ {{\mathbf{H}}_{1}},\cdots ,{{\mathbf{H}}_{m}} \right]$, $ {{\mathbf{W}}_{{{h}}}^m } \equiv [{{\mathbf{W}}_{{{h}_{1}}}} \cdots {{\mathbf{W}}_{{{h}_{m}}}}]$, $ {{\mathbf{\beta }}_{{{h}}}^m} \equiv  [{{\mathbf{\beta }}_{{{h}_{1}}}} \cdots {{\mathbf{\beta }}_{{{h}_{m}}}}]$;
\State \textbf{return}   ${{\mathbf{Z}}^{n}}$, ${{\mathbf{H}}^{m}}$, ${{\mathbf{W}}_{{{e}}}^n }$, ${{\mathbf{\beta }}_{{{e}}}^n} $, ${{\mathbf{W}}_{{{h}}}^m }$, ${{\mathbf{\beta }}_{{{h}}}^m}$
\EndFunction
\end{algorithmic}
\end{algorithm}

 The  $n$ groups of feature nodes ${{\mathbf{Z}}^{n}}$
 and the  $m$ groups of enhancement nodes ${{\mathbf{H}}^{m}}$
 form the expanded input matrix
 \begin{equation}\label{Anm2HZ438015}
\mathbf{A}^{n,m}=\left[ {{\mathbf{Z}}^{n}}|{{\mathbf{H}}^{m}} \right],
\end{equation}
where the superscripts ${n,m}$ in  $\mathbf{A}^{n,m}$ denote the numbers of feature and enhancement node  groups, respectively.
Finally,
 the connections of
 all the nodes
 are
  fed into the output by
\begin{equation}\label{Y2ZiHjWj948934}
\mathbf{\hat{Y}}=\mathbf{A}^{n,m} {{\mathbf{W}}^{n,m}},
\end{equation}
where the output weight ${{\mathbf{W}}^{n,m}} \in {\Re ^{k \times c}} $, the output ${\mathbf{\hat{Y}}} \in {\Re ^{l \times c}}$,
 and $c$  is  the size of the output.
 Assume that there are $k$ nodes totally. Then it can be seen that the expanded input matrix $\mathbf{A}^{n,m}\in {\Re ^{l \times k}}$  includes
$l$ training samples and  $k$ nodes.
 In this paper, sometimes we add the subscript to a matrix to indicate the number of nodes, e.g.,  $\mathbf{A}^{n,m}_k$  and   ${{\mathbf{W}}^{n,m}_k}$,
 or add the dotted subscript to a matrix to indicate the number of  training samples, e.g.,  $\mathbf{A}^{n,m}_{\ddot l}$  and   ${{\mathbf{W}}^{n,m}_{\ddot l}}$.
 Moreover, sometimes we also omit the superscripts for simplicity. For example, we may write
 $\mathbf{A}^{n,m}_k$,     ${{\mathbf{W}}^{n,m}_k}$, $\mathbf{A}^{n,m}_{\ddot l}$  and   ${{\mathbf{W}}^{n,m}_{\ddot l}}$ as
 $\mathbf{A}_k$,     ${{\mathbf{W}}}_k$, $\mathbf{A}_{\ddot l}$  and   ${{\mathbf{W}}}_{\ddot l}$, respectively.


 \subsection{Generalized Inverse Solution and Ridge Solution for BLS}


 The least-square solution~\cite{27_ref_BL_trans_paper} of (\ref{Y2ZiHjWj948934}) is
  the generalized inverse solution~\cite{best_ridge_inv_paper213}
\begin{equation}\label{W2AinvY989565}
{{\mathbf{W}}}=\mathbf{A}^{+} \mathbf{Y},
\end{equation}
where the output label $\mathbf{Y}\in {\Re ^{l \times c}}$ corresponds to
the input $\mathbf{X}$, and the generalized inverse $\mathbf{A}^{+} \in {\Re ^{k \times l}} $
satisfies
 \begin{equation}\label{Ainv2AtAinvAt9096}
{{\bf{A}}^{+ }}={{(\mathbf{A}^{T}\mathbf{A})}^{-1}}\mathbf{A}^{T}.
 \end{equation}
 The BLS algorithm  in \cite{BL_trans_paper}
computes
 the generalized inverse $\mathbf{A}^{+ }$   by
 \begin{equation}\label{AinvLimNumda0AAiA1221}
 \mathbf{A}_{{}}^{+ }=\underset{\lambda \to 0}{\mathop{\lim }}\,{{(\mathbf{A}_{{}}^{T}\mathbf{A}+\lambda \mathbf{I})}^{-1}}\mathbf{A}_{{}}^{T},
\end{equation}
which is  the ridge regression approximation~\cite{BL_trans_paper} of the generalized
inverse (\ref{Ainv2AtAinvAt9096}).


 It is required to assume the ridge parameter $\lambda \to 0$ in the original BLS algorithm  \cite{BL_trans_paper}
 that is based on the ridge regression approximation of the generalized
inverse, i.e., (\ref{AinvLimNumda0AAiA1221}).
To achieve a better generalization performance,
instead of (\ref{AinvLimNumda0AAiA1221}),
 the BLS algorithms
  proposed in   \cite{mybrief2onBLInputs2019}  and \cite{mybrief1onBLNodes2019}  adopt
 the ridge solution~\cite{best_ridge_inv_paper213}
 \begin{equation}\label{xWbarMN2AbarYYa1341}
{\mathbf{\tilde W}}={{\bf{A}}^{\dagger }}{{\mathbf{Y}}},
 \end{equation}
where ${{\bf{A}}^{\dagger }}$ is  the ridge inverse~\cite{best_ridge_inv_paper213}  of ${\bf{A}}$, i.e.,
\begin{equation}\label{xAmnAmnTAmnIAmnT231413}
{{\bf{A}}^{\dagger }}={{\left( {{\bf{A}}^{T}}{\bf{A}}+\lambda \mathbf{I} \right)}^{-1}}{{\bf{A}}^{T}}.
\end{equation}
Accordingly in     \cite{mybrief2onBLInputs2019} and \cite{mybrief1onBLNodes2019},  the assumption of  $\lambda \to 0$ (for
  the generalized inverse with the ridge regression approximation in the existing BLS) is no longer required, and $\lambda$ can be any positive real number.

In the original BLS~\cite{BL_trans_paper}, the output weights are updated
 easily
 for any number of new added inputs or nodes, by  computing the
generalized inverse of those added  inputs or nodes in just one iteration.
The efficient  incremental BLS algorithms on added inputs and nodes~\cite{mybrief2onBLInputs2019,mybrief1onBLNodes2019}
can obtain the  ridge solution for the updated output weights,
 which include the recursive BLS algorithm on added inputs and
  the  square-root BLS algorithm on added inputs and nodes~\footnote{As in \cite{mybrief2onBLInputs2019}, we follow the naming method
   in \cite{my_inv_chol_paper,TransSP2003Blast},  where the recursive algorithm updates the inverse matrix recursively, and the  square-root algorithm
    updates the square-root (including the Cholesky factor) of the inverse matrix.},
as will be introduced in the following two subsections.

\subsection{The Recursive ALgorithm and the Square-Root Algorithm for the Incremental BLS on Added Inputs}

The BLS includes the incremental learning for the additional input training samples.
When  encountering new input samples with the corresponding output labels,
the modeled BLS can be remodeled in an
incremental way without a complete retraining process.
It updates the output weights incrementally, without retraining the whole network from the beginning.

Denote the additional input training samples  as ${{\mathbf{\bar X}}_{\ddot  p}}$
with $p$ training samples.
The incremental $p$ samples for feature nodes
corresponding to ${{\mathbf{\bar X}}_{\ddot  p}}$
are computed by
  \begin{equation}\label{}
{{\mathbf{\bar Z}}_{i}}=\phi (\mathbf{\bar X}{{\mathbf{W}}_{{{e}_{i}}}}+{{\mathbf{\beta }}_{{{e}_{i}}}})
 \end{equation}
  for $i=1,2,\cdots,n$,
and concatenated into
  \begin{equation}\label{}
{{\mathbf{\bar Z}}^{n}}\equiv \left[ \begin{matrix}
   {{\mathbf{\bar Z}}_{1}} & \cdots  & {{\mathbf{\bar Z}}_{n}}  \\
\end{matrix} \right].
 \end{equation}
Then the incremental $p$ samples for enhancement nodes
are computed by
  \begin{equation}\label{}
{{\mathbf{\bar H}}_{j}}=\xi ({{\mathbf{\bar Z}}^{n}}{{\mathbf{W}}_{{{h}_{j}}}}+{{\mathbf{\beta }}_{{{h}_{j}}}})
 \end{equation}
  for $j=1,2,\cdots,m$,
 and concatenated into
   \begin{equation}\label{}
{{\mathbf{\bar H}}^{m}}\equiv \left[ {{\mathbf{\bar H}}_{1}},\cdots, {{\mathbf{\bar H}}_{m}} \right].
 \end{equation}
The above described procedure to add inputs in BLS is summarized in \textbf{Algorithm 2} by the function
  \begin{equation}\label{}
 ({{\mathbf{\bar Z}}^{n}}, {{\mathbf{\bar H}}^{m}})=\emph{AddInputs}(\mathbf{\bar X}, {{\mathbf{W}}_{{{e}}}^n }, {{\mathbf{\beta }}_{{{e}}}^n} , {{\mathbf{W}}_{{{h}}}^m }, {{\mathbf{\beta }}_{{{h}}}^m}).
 \end{equation}

\begin{algorithm}
\caption{The Procedure to Add Inputs in BLS}\label{euclid}
\begin{algorithmic}[0]
\Function{${AddInputs}$}{$\mathbf{\bar X}$, ${{\mathbf{W}}_{{{e}}}^n }$, ${{\mathbf{\beta }}_{{{e}}}^n} $, ${{\mathbf{W}}_{{{h}}}^m }$, ${{\mathbf{\beta }}_{{{h}}}^m}$}
\State Compute ${{\mathbf{\bar Z}}_{i}}=\phi (\mathbf{\bar X}{{\mathbf{W}}_{{{e}_{i}}}}+{{\mathbf{\beta }}_{{{e}_{i}}}})$  for $i=1,2,\cdots,n$, where ${{\mathbf{W}}_{{{e}_{i}}}}$
and ${{\mathbf{\beta }}_{{{e}_{i}}}}$ are in ${{\mathbf{W}}_{{{e}}}^n }$ and ${{\mathbf{\beta }}_{{{e}}}^n} $, respectively; 
\State    Set     ${{\mathbf{\bar Z}}^{n}}\equiv \left[ \begin{matrix}
   {{\mathbf{\bar Z}}_{1}} & \cdots  & {{\mathbf{\bar Z}}_{n}}  \\
\end{matrix} \right]$;
\State  Compute ${{\mathbf{\bar H}}_{j}}=\xi ({{\mathbf{\bar Z}}^{n}}{{\mathbf{W}}_{{{h}_{j}}}}+{{\mathbf{\beta }}_{{{h}_{j}}}})$ for $j=1,2,\cdots,m$, where
${{\mathbf{W}}_{{{h}_{j}}}}$ and ${{\mathbf{\beta }}_{{{h}_{j}}}}$ are in ${{\mathbf{W}}_{{{h}}}^m }$ and ${{\mathbf{\beta }}_{{{h}}}^m}$, respectively; 
\State   Set ${{\mathbf{\bar H}}^{m}}\equiv \left[ {{\mathbf{\bar H}}_{1}},\cdots, {{\mathbf{\bar H}}_{m}} \right]$;
\State \textbf{return}   ${{\mathbf{\bar Z}}^{n}}$, ${{\mathbf{\bar H}}^{m}}$
\EndFunction
\end{algorithmic}
\end{algorithm}

The incremental samples for feature nodes and enhancement nodes
 form the expanded input matrix $\mathbf{\bar A}_{{{\ddot  p}}}\in {\Re ^{p \times k}}$,
  i.e.,
\begin{equation}\label{PaperEqu21AxZ23141}
\mathbf{\bar A}_{\ddot  p}^{{}}= \left[\mathbf{\bar Z}^{n} | {{\mathbf{\bar H}}^{m}}\right].
\end{equation}
Accordingly,
 the expanded input matrix $\mathbf{A}_{{{\ddot  l}}}$ is updated into
\begin{equation}\label{AxInputIncrease31232}
{{\bf{A}}_{\ddot l + \ddot p}} = {\left[ {\begin{array}{*{20}{c}}
{{\bf{A}}_{\ddot l}^T}&{{\bf{\bar A}}_{\ddot p}^T}
\end{array}} \right]^T},
\end{equation}
and the corresponding output labels $\mathbf{Y}_{{{\ddot  l}}}$ is updated into
 \begin{equation}\label{YincreaseApril1klsd23}
{{\bf{Y}}_{\ddot l + \ddot p}} = {\left[ {\begin{array}{*{20}{c}}
{{\bf{Y}}_{\ddot l}^T}&{{\bf{\bar Y}}_{\ddot p}^T}
\end{array}} \right]^T},
 \end{equation}
where
the output labels ${{\mathbf{\bar Y}}_{\ddot  p}} \in {\Re ^{p \times c}} $  correspond to
the added input  ${{\mathbf{\bar X}}_{\ddot  p}}$.
From ${\bf{A}}_{ {\ddot  l} +{\ddot  p}}$ and ${{\mathbf{Y}}_{ {\ddot  l} +{\ddot  p}}}$,
the output weights
can be computed by (\ref{W2AinvY989565}) or
(\ref{xWbarMN2AbarYYa1341}).

In each iteration, the original incremental BLS algorithm in \cite{BL_trans_paper} updates the generalized inverse  (with the ridge regression approximation)
$\mathbf{A}_{{\ddot  l}}^{+}$  into  $\mathbf{A}_{{ {\ddot  l} +{\ddot  p}}}^{+}$,
and then utilizes the submatrix in $\mathbf{A}_{{ {\ddot  l} +{\ddot  p}}}^{+}$ to
update the output weights ${\mathbf{W}}_{\ddot  l}   $  into  ${\mathbf{W}}_{ {\ddot  l} +{\ddot  p}}$.
The $l \times k$ expanded input
matrix ${\bf{A}}_{\ddot  l}   $  has more rows than columns, i.e., $l>k$, since usually there are more  training samples than nodes
 in the neural networks~\cite{27_ref_BL_trans_paper,BL_trans_paper}.
 Accordingly,  the
recursive and square-root BLS algorithms on added inputs
  proposed in  \cite{mybrief2onBLInputs2019}
 update  ${{\mathbf{Q}}_{\ddot  l}   } \in {\Re ^{k \times k}}$ defined by
 \begin{equation}\label{Qm1AAIdefine23213}
{{\mathbf{Q}}_{\ddot  l}   }={{\left( {{\bf{A}}_{\ddot  l}   ^{T}}{\bf{A}}_{\ddot  l}   +\lambda \mathbf{I} \right)}^{-1}}
\end{equation}
and the upper-triangular  ${{\mathbf{F}}_{\ddot  l   }} \in {\Re ^{k \times k}} $
 satisfying
\begin{equation}\label{Q2PiPiT9686954}
{\mathbf{F}}_{\ddot  l   }{{\mathbf{F}}_{\ddot  l   }^{T}}={{\mathbf{Q}}_{\ddot  l   }}={({{{{\bf{A}}_{\ddot  l   }^{T}}{\bf{A}}_{\ddot  l   }+\lambda \mathbf{I}}})^{-1}},
\end{equation}
respectively,
from which the output weights are computed.  Moreover, when there are more rows than columns in the newly added $p \times k$ input matrix ${\bf{\bar A}}_{\ddot  p}$ (corresponding to the added inputs  ${{\mathbf{\bar X}}_{\ddot  p}}$), i.e., $p > k$,
  the inverse of a sum of matrices~\cite{InverseSumofMatrix8312} is utilized by
   both BLS algorithms proposed in \cite{mybrief2onBLInputs2019}
   to  compute the intermediate variables by a smaller matrix inverse or inverse Cholesky factorization.

In the recursive BLS algorithm on added inputs~\cite{mybrief2onBLInputs2019},
   ${{\mathbf{Q}}_{\ddot  l   }}$ and ${\mathbf{\tilde W}}_{\ddot  l   }$ are updated into  ${{\mathbf{Q}}_{\ddot  l    + \ddot  p   }}$ and ${\mathbf{\tilde W}}_{\ddot  l    + \ddot  p   }$ by
\begin{subnumcases}{\label{BQWpSmall32kds32j}}
{{\mathbf{\tilde B}}}  = {{\mathbf{Q}}_{\ddot  l   }}\mathbf{\bar A}_{ \ddot  p   }^{T}{{(\mathbf{I}+\mathbf{\bar A}_{ \ddot  p   }^{{}}{{\mathbf{Q}}_{\ddot  l   }}\mathbf{\bar A}_{ \ddot  p   }^{T})}^{-1}} &  \label{BmatrixFromMyQ322a}\\
{{\mathbf{Q}}_{\ddot  l   + \ddot  p   }}={{\mathbf{Q}}_{\ddot  l   }}-{{\mathbf{\tilde B}}} \mathbf{\bar A}_{ \ddot  p   }^{{}}{{\mathbf{Q}}_{\ddot  l   }}  &  \label{BmatrixFromMyQ322aForQ}\\
{\mathbf{\tilde W}}_{\ddot  l    + \ddot  p   }=\mathbf{\tilde W}_{\ddot  l   }+ {{\mathbf{\tilde B}}}
  \left( {{\mathbf{\bar Y}}_{ \ddot  p   }}-\mathbf{\bar A}_{ \ddot  p   }^{{}}\mathbf{\tilde W}_{\ddot  l   } \right)  &  \label{xWmnIncreaseCompute86759Before}
\end{subnumcases}
when  $p<k$,
or
 by
\begin{subnumcases}{\label{QWpBig32148kd43}}
{{\mathbf{Q}}_{\ddot  l   + \ddot  p   }}={{(\mathbf{I}+{{\mathbf{Q}}_{\ddot  l   }}\mathbf{\bar A}_{ \ddot  p   }^{T} \mathbf{\bar A}_{ \ddot  p   }^{{}})}^{-1}}{{\mathbf{Q}}_{\ddot  l   }} &  \label{BmatrixFromMyQ322bForQ}\\
{\mathbf{\tilde W}}_{\ddot  l    + \ddot  p   }=\mathbf{\tilde W}_{\ddot  l   }+ {{\mathbf{Q}}_{\ddot  l   + \ddot  p   }} \mathbf{\bar A}_{ \ddot  p   }^{T}
  \left({{\mathbf{\bar Y}}_{ \ddot  p   }}-\mathbf{\bar A}_{ \ddot  p   }^{{}}\mathbf{\tilde W}_{\ddot  l   } \right)  &  \label{xWmnIncreaseCompute86759use231}
\end{subnumcases}
when $p \ge k$.
On the other hand,  the square-root BLS algorithm on added inputs~\cite{mybrief2onBLInputs2019}
computes the intermediate matrix $\mathbf{S} \in {\Re ^{k \times p}}$ by
\begin{equation}\label{K2AxLm94835}
\mathbf{S}=\mathbf{F}_{{{\ddot  l   }}}^{T}\mathbf{\bar A}_{ \ddot  p   }^{T},
\end{equation}
and updates ${{\mathbf{F}}_{{{\ddot  l   }}}}$ into ${{\mathbf{F}}_{{\ddot  l   }+\bar{\ddot  p}}}$
by
\begin{equation}\label{Lbig2LLwave59056}
{{\mathbf{F}}_{{\ddot  l   }+\bar{\ddot  p}}}={{\mathbf{F}}_{{{\ddot  l   }}}}\mathbf{V}.
\end{equation}
To compute the upper-triangular intermediate matrix
$\mathbf{V}$ in (\ref{Lbig2LLwave59056}) and update
${\mathbf{\tilde W}}_{\ddot  l   }$  into  ${\mathbf{\tilde W}}_{\ddot  l    + \ddot  p   }$,
the square-root BLS algorithm utilizes
\begin{subnumcases}{\label{VWpSmallThaNk32230}}
{\mathbf{V}} {\mathbf{V}}^{T}= \mathbf{I}- \mathbf{S} {{(\mathbf{I}+{{\mathbf{S}}^{T}}\mathbf{S})}^{-1}}{{\mathbf{S}}^{T}} &  \label{Lwave2IKK40425b}\\
{\mathbf{\tilde W}}_{\ddot  l    + \ddot  p   } = {{{\bf{\tilde W}}}_{\ddot  l   }} + {\mathbf{F}}_{\ddot  l   } \mathbf{S} {{(\mathbf{I}+\mathbf{S}^{T}\mathbf{S} )}^{-1}} ({{{\mathbf{\bar Y}}_{ \ddot  p   }} - \mathbf{\bar A}_{ \ddot  p   }^{}{{{\bf{\tilde W}}}_{\ddot  l   }}})   &  \label{WfromF3209da}
\end{subnumcases}
when  $p<k$,  or utilizes
\begin{subnumcases}{\label{VWpBigThaNk02kds32s}}
{\mathbf{V}} {\mathbf{V}}^{T} ={{(\mathbf{I}+\mathbf{S} {{\mathbf{S}}^{T}})}^{-1}} &  \label{Lwave2IKK40425a}\\
{\mathbf{\tilde W}}_{\ddot  l    + \ddot  p   } = {{{\bf{\tilde W}}}_{\ddot  l   }} + {{\bf{F}}_{\ddot  l    + \ddot  p   }}{\bf{F}}_{\ddot  l    + \ddot  p   }^T \mathbf{\bar A}_{ \ddot  p   }^T ({{{\mathbf{\bar Y}}_{ \ddot  p   }} - \mathbf{\bar A}_{ \ddot  p   }^{}{{{\bf{\tilde W}}}_{\ddot  l   }}})  &  \label{WfromF3209db}
\end{subnumcases}
when $p \ge k$.

\subsection{The Square-Root Algorithm for the Incremental BLS on Added Nodes}


The BLS also includes the incremental learning for the additional nodes. When the network
does not reach the desired
accuracy,  additional
nodes can be inserted to achieve a better performance, by fast
remodeling in broad expansion without a retraining process.
Denote the additional  $q$   nodes
as
$\mathbf{\bar A}_{q}$
with $q$ columns.
Then  the expanded input matrix should be updated into
\begin{equation}\label{Anp2AnH954734}
\mathbf{A}_{k+q}^{{}}=\left[ \mathbf{A}_{k}^{{}}| \mathbf{\bar A}_{q}  \right].
 \end{equation}

The square-root BLS algorithm on added nodes in \cite{mybrief1onBLNodes2019} defines
 \begin{equation}\label{R_define12321numda}
 {{\bf{R}}_k}={{\bf{A}}_k^T}{{\bf{A}}_k}+ \lambda {\bf{I}},
 \end{equation}
 and then   ${{\bf{R}}_{k+q}}$ can be  written
 as
   a $2 \times 2$ block Hermitian matrix
 \begin{equation}\label{R_def_perhaps_No_R1}{{\bf{R}}_{k+q}}=\left[ \begin{matrix}
   {{\bf{R}}_k} & {\mathbf{\tilde R}}_{k,q}  \\
  {\mathbf{\tilde R}}_{k,q}^{T} & {{\bf{\bar R}}_q}  \\
\end{matrix} \right],
\end{equation}
where ${\mathbf{\tilde R}}_{k,q}\in {\Re ^{k \times q}}$ and ${{\bf{\bar R}}_q} \in {\Re ^{q \times q}}$
 satisfy
 \begin{subnumcases}{\label{EFdefine13214ZF13ab}}
{\mathbf{\tilde R}}_{k,q} ={{\bf{A}}_k^T}\mathbf{\bar A}_{q} &  \label{EFdefine13214ZF13a}\\
{{\bf{\bar R}}_q} =\mathbf{\bar A}_{q}^{T}\mathbf{\bar A}_{q}+\lambda \mathbf{I}. & \label{EFdefine13214ZF13b}
\end{subnumcases}

An efficient inverse Cholesky factorization was proposed in \cite{my_inv_chol_paper}
to compute  the upper-triangular inverse Cholesky factor of
 ${\bf{R}}_k$, i.e.,
 ${{\bf{F}}_k}$
 satisfying
\begin{equation}\label{L_m_def12431before}
{{\bf{F}}_k}{{\bf{F}}_k^{T}}={{\bf{R}}_k^{-1}}=({{\bf{A}}_k^T}{{\bf{A}}_k}+ \lambda {\bf{I}})^{-1}.
\end{equation}
Then in \cite{mybrief1onBLNodes2019},
the inverse Cholesky factorization~\cite{my_inv_chol_paper}
was extended
to  compute  the upper-triangular inverse Cholesky factor of
the $2 \times 2$ block matrix
 ${{\bf{R}}_{k+q}}$  described in  (\ref{R_def_perhaps_No_R1}).
The block inverse Cholesky factorization  proposed   in  \cite{mybrief1onBLNodes2019}
obtain  ${{\bf{F}}_{k+q}}$
from ${{\bf{F}}_k}$ by
\begin{equation}\label{L_big_BLK_def1}{{\bf{F}}_{k+q}}=\left[ \begin{matrix}
   {{\bf{F}}_k} & {{\bf{\tilde F}}_{k,q}}  \\
   \mathbf{0} & {{\bf{\bar F}}_q}  \\
\end{matrix} \right],
\end{equation}
where
 ${{\bf{\bar F}}_q} \in {\Re ^{q \times q}}$
 and
 ${{\bf{\tilde F}}_{k,q}}\in {\Re ^{k \times q}}$
are computed by
\begin{subnumcases}{\label{ZF_def_L_2_items3ab}}
{{{\bf{\bar F}}_q}}{{\bf{\bar F}}_q^{T}}=\left({{\bf{\bar R}}_q}- {\mathbf{\tilde R}}_{k,q}^{T}{{\bf{F}}_k}{{\bf{F}}_k^{T}}{\mathbf{\tilde R}}_{k,q}\right)^{-1} &  \label{ZF_def_L_2_items3a}\\
 {{\bf{\tilde F}}_{k,q}}=-{{\bf{F}}_k}{{\bf{F}}_k^{T}}{\mathbf{\tilde R}}_{k,q}{{{\bf{\bar F}}_q}}. & \label{ZF_def_L_2_items3b}
\end{subnumcases}
Notice that ${{{\bf{\bar F}}_q}}$ in (\ref{ZF_def_L_2_items3a})
  is
 the upper-triangular
 inverse Cholesky factor
  of
${{\bf{\bar R}}_q}- {\mathbf{\tilde R}}_{k,q}^{T}{{\bf{F}}_k}{{\bf{F}}_k^{T}}{\mathbf{\tilde R}}_{k,q}$,
 which  can be computed by the inverse Cholesky factorization~\cite{my_inv_chol_paper},
or by inverting and transposing the lower-triangular Cholesky
factor.

The above-described block inverse Cholesky factorization has been  applied to
develop the the square-root BLS algorithm on added nodes
  in \cite{mybrief1onBLNodes2019}.
The BLS algorithm in \cite{mybrief1onBLNodes2019} computes
the initial ${{\bf{F}}_k}$
from ${{\bf{A}}_k}$ by (\ref{L_m_def12431before}),
and then obtains
 ${{\bf{F}}_{k+q}}$
from ${{\bf{F}}_k}$ by
(\ref{EFdefine13214ZF13ab}),
(\ref{ZF_def_L_2_items3ab})
and
(\ref{L_big_BLK_def1}),
and utilizes the sub-matrices
${{\bf{\bar F}}_q}$
and
 ${{\bf{\tilde F}}_{k,q}}$  in ${{\bf{F}}_{k+q}}$ (defined by   (\ref{L_big_BLK_def1}))
  to compute the ridge solution for the output weights by
\begin{equation}\label{W2AMHEWNH3495}{ {\bf{\tilde W}}_{k+q} }=\left[ \begin{matrix}
   {{\bf{\tilde W}}_k}+{{\bf{\tilde F}}_{k,q}}{{\bf{\bar F}}_q^{T}}\left( \mathbf{\bar A}_{q}^{T}\mathbf{Y}-{\mathbf{\tilde R}}_{k,q}^{T}{{\bf{\tilde W}}_k} \right)  \\
   {{\bf{\bar F}}_q} {{\bf{\bar F}}_q^{T}}\left( \mathbf{\bar A}_{q}^{T}\mathbf{Y}-{\mathbf{\tilde R}}_{k,q}^{T}{{\bf{\tilde W}}_k} \right)  \\
\end{matrix} \right],
\end{equation}
where the initial ${ {\bf{\tilde W}}_{k} }$ is computed by
\begin{equation}\label{W2AMHEWNH3495initial}
{ {\bf{\tilde W}}_{k} }= {{\bf{F}}_k}{{\bf{F}}_k^{T}}\mathbf{A}_{{k}}^{T} {{\mathbf{Y}}}.
\end{equation}



\section{Low-Memory Implementations of the Existing BLS Algorithms by Processing a Batch of Inputs or Nodes in Each Recursion}


In \cite{BLSLowMemTNNLS2020dec}, it has been mentioned that
out-of-memory problems occur in the original BLS
due to inverting  too large matrices, and
are easily faced in standard desktop PCs
(Intel Core i5 Quad-Core with 8 GB DDR4, and BLS implemented
in MATLAB) when $l$, $k$ and $c$ (i.e., the numbers of samples, nodes and outputs)
satisfy  $l > 100 000$, $k > 10 000$ and $c > 100$, respectively.
Then to avoid the need to invert  too large matrices,
the hybrid recursive BLS implementation proposed in \cite{BLSLowMemTNNLS2020dec}
divides $l$ training samples into small batches and processes only one batch in each recursion.
When there are $b$ training samples in each small batch,
only a $b \times b$ matrix inversion is required in each recursion.

In this section, we will propose the  low-memory implementations for
the recursive and  square-root BLS Algorithms on added inputs
proposed in \cite{mybrief2onBLInputs2019}, and the square-root BLS Algorithm on added nodes
proposed in \cite{mybrief1onBLNodes2019}. The proposed implementations
   also
 process only one batch of $b$ inputs or nodes  in each recursion by
 a $b \times b$ matrix inversion or inverse Cholesky factorization, to avoid the inversion of too large matrices
 that causes out-of-memory problems.

\subsection{The Proposed Low-Memory Implementation of the Recursive BLS Algorithm on Added Inputs}

%
%

To process a batch of $b$ samples
(by only a $b \times b$ matrix inversion or inverse Cholesky factorization) in each recursion,
 let us divide the expanded input matrix $\mathbf{A}_{{{\ddot  l}}}$ with $l$ training samples
 into $l/b$ batches and  each batch includes $b$ samples.
   Accordingly,  we have
 \begin{equation}\label{Al2AlbAb1b20923ds23ds}
{{\bf{A}}_{\ddot l}} =  {\left[ {\begin{array}{*{20}{c}}
{{\bf{A}}_{{\ddot b}_b}^T}&{{\bf{A}}_{{\ddot b}_{2b}}^T}& \cdots  & {{\bf{A}}_{{\ddot b}_{i}}^T}  & \cdots  & {{\bf{A}}_{{\ddot b}_{l}}^T}
\end{array}} \right]^T},
 \end{equation}
 where ${{\bf{A}}_{{\ddot b}_{i}}}$ ($i=b,2b,\cdots,(l/b)b$) denotes the $b$ rows of $\mathbf{A}_{{\ddot  l}}$ from row $i-b+1$ to row $i$.
Accordingly,  we can
process ${{\bf{A}}_{{\ddot b}_{i}}}$
 to
 update ${{\mathbf{Q}}_{\ddot i- \ddot b}}$
 and $\mathbf{\tilde W}_{\ddot i- \ddot b}$ into
 ${{\mathbf{Q}}_{\ddot i}}$  and  ${\mathbf{\tilde W}}_{\ddot i}$, respectively, by
 simply setting
 $ l= i-  b$
 and
 $ p =  b$
    in (\ref{BQWpSmall32kds32j}) to obtain
\begin{subnumcases}{\label{BQWpSmall32kds32jnb}}
{{\mathbf{\tilde B}}}  = {{\mathbf{Q}}_{\ddot i- \ddot b}}\mathbf{A}_{ {{\ddot b}_i}  }^{T}{{(\mathbf{I}+\mathbf{A}_{ {{\ddot b}_i}  }^{{}}{{\mathbf{Q}}_{\ddot i- \ddot b}}\mathbf{A}_{ {{\ddot b}_i}  }^{T})}^{-1}} &  \label{BmatrixFromMyQ322anb}\\
{{\mathbf{Q}}_{\ddot i}  }={{\mathbf{Q}}_{\ddot i- \ddot b}}-{{\mathbf{\tilde B}}} \mathbf{A}_{ {{\ddot b}_i}  }^{{}}{{\mathbf{Q}}_{\ddot i- \ddot b}}  &  \label{BmatrixFromMyQ322aForQnb}\\
{\mathbf{\tilde W}}_{\ddot i}  =\mathbf{\tilde W}_{\ddot i- \ddot b}+ {{\mathbf{\tilde B}}}
  \left( {{\mathbf{Y}}_{ {{\ddot b}_i}  }}-\mathbf{A}_{ {{\ddot b}_i}  }^{{}}\mathbf{\tilde W}_{\ddot i- \ddot b} \right),  &  \label{xWmnIncreaseCompute86759Beforenb}
\end{subnumcases}
where ${{\mathbf{Y}}_{ {{\ddot b}_i}  }}$
 denotes
the $b$ rows of $\mathbf{Y}_{{\ddot  l}}$ from row $i-b+1$ to row $i$.

When $i=b$,
${{\bf{Q}}_{\ddot 0}}$ and   ${\mathbf{\tilde W}}_{\ddot 0}$ are
required to
  compute ${{\mathbf{Q}}_{\ddot b}}$  and  ${\mathbf{\tilde W}}_{\ddot b}$ by (\ref{BQWpSmall32kds32jnb}).
To deduce  ${{\bf{Q}}_{\ddot 0}}$ and   ${\mathbf{\tilde W}}_{\ddot 0}$,
set $ l = 0$ to write
(\ref{Qm1AAIdefine23213})
and
(\ref{xWbarMN2AbarYYa1341})
 as
    \begin{equation}\label{QlamdaI201dksQ0}
{{\bf{Q}}_{\ddot 0}} = {\left( {\lambda {\bf{I}}} \right)^{ - 1}} = ({1}/{\lambda }){\bf{I}}
 \end{equation}
 and
 \begin{equation}\label{xWbarMN2AbarYYa1341Empty}
{\mathbf{\tilde W}}_{\ddot 0}=\pmb{\Phi},
 \end{equation}
 respectively,
 where $\pmb{\Phi}$ denotes the empty matrix.
Then
(\ref{QlamdaI201dksQ0})
and
(\ref{xWbarMN2AbarYYa1341Empty})
can be substituted
into
(\ref{BmatrixFromMyQ322anb}),
(\ref{BmatrixFromMyQ322aForQnb}) and
(\ref{xWmnIncreaseCompute86759Beforenb})
with $i=b$
to obtain
  \begin{equation}\label{Step2BmatrixFromMyQ322a}
{{\mathbf{\tilde B}}}  = \frac{\mathbf{A}_{ {{\ddot b}_b}  }^{T}}{\lambda} {{\left(\mathbf{I}+\frac{\mathbf{A}_{ {{\ddot b}_b}  } \mathbf{A}_{ {{\ddot b}_b}  }^{T} }{\lambda}\right)}^{-1}}
 = \mathbf{A}_{ {{\ddot b}_b}  }^{T}{{(\mathbf{A}_{ {{\ddot b}_b}  }\mathbf{A}_{ {{\ddot b}_b}  }^{T} +{\lambda }\mathbf{I})}^{-1}},
 \end{equation}
  \begin{equation}\label{Step2BmatrixFromMyQ322aForQ}
{{\mathbf{Q}}_{\ddot b}  }=({1}/{\lambda }){\bf{I}}-{{\mathbf{\tilde B}}} \mathbf{A}_{ {{\ddot b}_b}  }^{{}}({1}/{\lambda }){\bf{I}}
= ({\bf{I}}-{{\mathbf{\tilde B}}} \mathbf{A}_{ {{\ddot b}_b}  })/{\lambda }
 \end{equation}
and
     \begin{equation}\label{Step2xWmnIncreaseCompute86759Before}
{\mathbf{\tilde W}}_{{\ddot b}}  =\pmb{\Phi}+ {{\mathbf{\tilde B}}}
  \left( {{\mathbf{Y}}_{ {{\ddot b}_b}  }}-\mathbf{A}_{ {{\ddot b}_b}  }^{{}}\pmb{\Phi} \right)
  = {{\mathbf{\tilde B}}} {{\mathbf{Y}}_{ {{\ddot b}_b}  }},
 \end{equation}
 respectively,
   which can be
written as
  \begin{subnumcases}{\label{StepLastBQWpSmall32kds32j}}
{{\mathbf{\tilde B}}}  = \mathbf{A}_{ {{\ddot b}_b}  }^{T}{{(\mathbf{A}_{ {{\ddot b}_b}  }\mathbf{A}_{ {{\ddot b}_b}  }^{T} +{\lambda }\mathbf{I})}^{-1}} &  \label{StepLastBmatrixFromMyQ322a}\\
{{\mathbf{Q}}_{\ddot b}} = ({\bf{I}}-{{\mathbf{\tilde B}}} \mathbf{A}_{ {{\ddot b}_b}  })/{\lambda }  &  \label{StepLastBmatrixFromMyQ322aForQ}\\
{\mathbf{\tilde W}}_{\ddot b} = {{\mathbf{\tilde B}}} {{\mathbf{Y}}_{ {{\ddot b}_b}  }}.  &  \label{StepLastxWmnIncreaseCompute86759Before}
\end{subnumcases}

Now we can compute ${{\mathbf{Q}}_{\ddot b}}$ and ${\mathbf{\tilde W}}_{\ddot b}$  by  (\ref{StepLastBQWpSmall32kds32j}) firstly, and then
compute (\ref{BQWpSmall32kds32jnb}) recursively for $i=2b,3b,\cdots,(l/b)b$,
to obtain
 ${\mathbf{Q}}_{\ddot  l}  $ and $ {\mathbf{\tilde W}}_{\ddot  l}   $.
When  new input samples are encountered,
we can  obtain ${{\mathbf{Q}}_{\ddot l+\ddot p}}$  and  ${\mathbf{\tilde W}}_{\ddot l+\ddot p}$ by
computing
 (\ref{BQWpSmall32kds32jnb}) iteratively for $i=l+b,l+2b,\cdots,l+(p/b)b$,
 where $\mathbf{A}_{ {{\ddot b}_i}  }$
  denotes the $b$ rows of ${\bf{A}}_{ {\ddot  l} +{\ddot  p}}$ (described in (\ref{AxInputIncrease31232})) from row $i-b+1$ to row $i$.


We summarize
the above-described low-memory implementation of
the recursive BLS algorithm for added inputs~\cite{mybrief2onBLInputs2019}
in the following \textbf{Algorithm 3}, where the functions \emph{Initialize}
and \emph{AddInputs} are described in \textbf{Algorithm 1} and \textbf{Algorithm 2},
respectively.


\begin{algorithm}
\caption{:~\bf The Low-Memory Implementation of the Recursive BLS Algorithm on Added Inputs:  Computation of Output Weights and  Increment of  New Inputs}
\begin{algorithmic}[1]
\Require   Inputs $\mathbf{X}_{{\ddot  l}}$ with labels $\mathbf{Y}_{{\ddot  l}}$, added inputs with labels;
\Ensure Output weights: the ridge solution $\mathbf{\tilde W}$;
\State  Get  (${{\mathbf{Z}}^{n}}$, ${{\mathbf{H}}^{m}}$, ${{\mathbf{W}}_{{{e}}}^n }$, ${{\mathbf{\beta }}_{{{e}}}^n} $, ${{\mathbf{W}}_{{{h}}}^m }$, ${{\mathbf{\beta }}_{{{h}}}^m}$)=
\emph{Initialize} ($\mathbf{X}_{{\ddot  l}}$);
\State   Set $\mathbf{A}_{{\ddot  l}}=\left[ {{\mathbf{Z}}^{n}}|{{\mathbf{H}}^{m}} \right]$;
\State   Utilize $\mathbf{A}_{{\ddot  l}}$ and $\mathbf{Y}_{{\ddot  l}}$  to compute ${{\mathbf{Q}}_{\ddot b}}$ and ${\mathbf{\tilde W}}_{\ddot b}$  by  (\ref{StepLastBQWpSmall32kds32j}), and then obtain  ${\mathbf{Q}}_{\ddot  l}  $ and $ {\mathbf{\tilde W}}_{\ddot  l}   $ by computing (\ref{BQWpSmall32kds32jnb}) iteratively  for $i=2b,3b,\cdots,(l/b)b$;
\While{\emph{New inputs $\mathbf{\bar X}_{{\ddot  p}}$ and labels $\mathbf{\bar Y}_{{\ddot  p}}$ are added}}
\State  Get   (${{\mathbf{\bar Z}}_{{\ddot  p}}^{n}}$, ${{\mathbf{\bar H}}_{{\ddot  p}}^{m}}$)=\emph{AddInputs}($\mathbf{\bar X}_{{\ddot  p}}$, ${{\mathbf{W}}_{{{e}}}^n }$, ${{\mathbf{\beta }}_{{{e}}}^n} $, ${{\mathbf{W}}_{{{h}}}^m }$, ${{\mathbf{\beta }}_{{{h}}}^m}$);
\State   Set $\mathbf{\bar A}_{{{\ddot  p}}}=\left[ {{\mathbf{\bar Z}}_{{\ddot  p}}^{n}}|{{\mathbf{\bar H}}_{{\ddot  p}}^{m}} \right]$;
\State   Utilize $\mathbf{\bar A}_{{{\ddot  p}}}$ and $\mathbf{\bar Y}_{{\ddot  p}}$  to update
${{\mathbf{Q}}_{\ddot l}}$  and  ${\mathbf{\tilde W}}_{\ddot l}$
into  ${{\mathbf{Q}}_{\ddot l+\ddot p}}$  and  ${\mathbf{\tilde W}}_{\ddot l+\ddot p}$, respectively, by
computing  (\ref{BQWpSmall32kds32jnb}) iteratively for $i=l+b,l+2b,\cdots,l+(p/b)b$;
\State Set ${{\bf{A}}_{\ddot l + \ddot p}} = {\left[ {\begin{array}{*{20}{c}}
{{\bf{A}}_{\ddot l}^T}&{{\bf{\bar A}}_{\ddot p}^T}
\end{array}} \right]^T}$ and $l=l+p$;
\EndWhile
\State Set output weights $\mathbf{\tilde W}={\mathbf{\tilde W}}_{\ddot  l}   $;
\end{algorithmic}
\end{algorithm}

\subsection{The Proposed Low-Memory Implementation of the Square-Root BLS Algorithm on Added Inputs}

  As in the last subsection,
 let us simply set $l=\ddot i- \ddot b$ and  $ p = b$   in
 (\ref{K2AxLm94835}),
(\ref{Lwave2IKK40425b}),
(\ref{WfromF3209da}) and
(\ref{Lbig2LLwave59056})
  to obtain
  \begin{subnumcases}{\label{LowMemChol3209dzsAll}}
\mathbf{S}=\mathbf{F}_{\ddot i- \ddot b}^{T}  \mathbf{A}_{ {{\ddot b}_i}  }^{T} &  \label{LowMemChol3209dzs1}\\
 {\mathbf{V}} {\mathbf{V}}^{T} = \mathbf{I}- \mathbf{S} {{(\mathbf{I}+{{\mathbf{S}}^{T}}\mathbf{S})}^{-1}}{{\mathbf{S}}^{T}}  &  \label{LowMemChol3209dzs2}\\
\begin{array}{r}
{{{\bf{\tilde W}}}_{\ddot i}  } = {{{\bf{\tilde W}}}_{\ddot i- \ddot b}} + {{\bf{F}}_{\ddot i- \ddot b}}{\bf{S}} {\left( {{\bf{I}} + {{\bf{S}}^T}{\bf{S}}} \right)^{ - 1}} \times  \\
 \left( {{{\bf{Y}}_{{{\ddot b}_i} }} - {{\bf{A}}_{{{\ddot b}_i} }}{{{\bf{\tilde W}}}_{\ddot i- \ddot b}}} \right)
\end{array}    &  \label{LowMemChol3209dzs3}\\
 {{\mathbf{F}}_{\ddot i }}={{\mathbf{F}}_{\ddot i- \ddot b}}\mathbf{V},  &  \label{LowMemChol3209dzs4}
\end{subnumcases}
which
process ${{\bf{A}}_{{{\ddot b}_i}}}$ in the $i^{th}$ ($i=1,2,\cdots,l/b$) recursion to
update ${{\mathbf{F}}_{\ddot i- \ddot b}}$ and $\mathbf{\tilde W}_{\ddot i- \ddot b}$ into
 ${{\mathbf{F}}_{\ddot i}  }$  and  ${\mathbf{\tilde W}}_{\ddot i}  $, respectively.

When $i=b$,  ${{\mathbf{F}}_{\ddot b}}$  and  ${\mathbf{\tilde W}}_{\ddot b}$  are computed from
${{\bf{F}}_{\ddot 0}}$  and ${\mathbf{\tilde W}}_{\ddot 0}$ by  (\ref{LowMemChol3209dzsAll}).
To deduce ${{\bf{F}}_{\ddot 0}}$,
let us
 substitute (\ref{Q2PiPiT9686954})
 into (\ref{QlamdaI201dksQ0}) to obtain
${\mathbf{F}}_{\ddot 0}{{\mathbf{F}}_{\ddot 0}^{T}} = \frac{1}{\lambda}{\bf{I}}$,
from which
 we can obtain a ${\mathbf{F}}_{\ddot 0}$, i.e.,
\begin{equation}\label{F0LamdaI3293232}
{\mathbf{F}}_{\ddot 0} = \frac{1}{\sqrt{ \lambda}}{\bf{I}}.
 \end{equation}
  To obtain ${{\mathbf{F}}_{\ddot b}}$,
 the above
 (\ref{F0LamdaI3293232})  is substituted  into
(\ref{LowMemChol3209dzs1}) and (\ref{LowMemChol3209dzs4}) (with $i=b$)
 to obtain
  \begin{equation}\label{S2LamdaAp320923sd}
\mathbf{S}=\frac{\mathbf{A}_{ {{\ddot b}_b}  }^{T}}{\sqrt{ \lambda}},
 \end{equation}
 and
 \begin{equation}\label{F2LamdaV0320943sd}
 \mathbf{V} = \sqrt{\lambda} {{\mathbf{F}}_{\ddot b}},
 \end{equation}
 respectively,
which are then substituted
    into (\ref{LowMemChol3209dzs2})
 to obtain
 ${\sqrt{ \lambda}} {{\mathbf{F}}_{\ddot b}} {\sqrt{ \lambda}} {{\mathbf{F}}_{\ddot b}^{T}} = \mathbf{I}-
  \frac{\mathbf{A}_{ {{\ddot b}_b}  }^{T}}{\sqrt{ \lambda}}
  { \left(\mathbf{I}+
     \frac{\mathbf{A}_{ {{\ddot b}_b}  }}{\sqrt{ \lambda}}
     \frac{\mathbf{A}_{ {{\ddot b}_b}  }^{T}}{\sqrt{ \lambda}}\right)^{-1}  }   \frac{\mathbf{A}_{ {{\ddot b}_b}  }}{\sqrt{ \lambda}}$, i.e.,
 \begin{equation}\label{FbFbLamda30929dks23cxd}
{\mathbf{F}}_{\ddot b}{{\mathbf{F}}_{\ddot b}^{T}} = \frac{1}{\lambda}{\bf{I}}- \frac{1}{\lambda} \mathbf{A}_{ {{\ddot b}_b}  }^{T}{{(\mathbf{A}_{ {{\ddot b}_b}  }\mathbf{A}_{ {{\ddot b}_b}  }^{T} +{\lambda }\mathbf{I})}^{-1}}  \mathbf{A}_{ {{\ddot b}_b}  }.
 \end{equation}
Finally,
to obtain ${\mathbf{\tilde W}}_{\ddot b}$, let us
 substitute  (\ref{F0LamdaI3293232}),  (\ref{S2LamdaAp320923sd}) and (\ref{xWbarMN2AbarYYa1341Empty}) (i.e., ${\mathbf{\tilde W}}_{\ddot 0}=\pmb{\Phi}$)
 into  (\ref{LowMemChol3209dzs3}) (with $i=b$)
 to obtain
  ${{\bf{\tilde W}}_{\ddot b}} =\pmb{\Phi} +\frac{1}{\sqrt{ \lambda}}  \frac{\mathbf{A}_{ {{\ddot b}_b}  }^{T}}{\sqrt{ \lambda}}  {\left( {{\bf{I}} + \frac{\mathbf{A}_{ {{\ddot b}_b}  }}{\sqrt{ \lambda}} \frac{\mathbf{A}_{ {{\ddot b}_b}  }^{T}}{\sqrt{ \lambda}} } \right)^{ - 1}}\left( {{{\bf{Y}}_{{{\ddot b}_b} }} - {{\bf{A}}_{{{\ddot b}_b} }}  \pmb{\Phi}} \right)$, i.e.,
\begin{equation}\label{Wb2Ab0239ald2320akc}
{\mathbf{\tilde W}}_{\ddot b} = \mathbf{A}_{ {{\ddot b}_b}  }^{T}{{(\mathbf{A}_{ {{\ddot b}_b}  }\mathbf{A}_{ {{\ddot b}_b}  }^{T} +{\lambda }\mathbf{I})}^{-1}}     {{\mathbf{Y}}_{ {{\ddot b}_b}  }}.
 \end{equation}



 Now we can compute  the upper-triangular ${{\mathbf{F}}_{\ddot b}}$ by (\ref{FbFbLamda30929dks23cxd}), and compute
     ${\mathbf{\tilde W}}_{\ddot b}$ by (\ref{Wb2Ab0239ald2320akc}).
   Then we
compute (\ref{LowMemChol3209dzsAll}) iteratively for $i=2b,3b,\cdots,(l/b)b$,
to obtain
 ${{\mathbf{F}}_{\ddot  l}   } $ and $ {\mathbf{\tilde W}}_{\ddot  l}   $.
When  new input samples are encountered,
we can compute
 (\ref{LowMemChol3209dzsAll}) iteratively for $i=l+b,l+2b,\cdots,l+(p/b)b$,
 where $\mathbf{A}_{ {{\ddot b}_i}  }$
  denotes the $b$ rows of ${\bf{A}}_{ {\ddot  l} +{\ddot  p}}$
   from row $i-b+1$ to row $i$,
to obtain ${{\mathbf{F}}_{\ddot l+\ddot p}}$  and  ${\mathbf{\tilde W}}_{\ddot l+\ddot p}$ finally.


In  \textbf{Algorithm 4}, we summarize
the above-described low-memory implementation of
the square-root BLS algorithm for added inputs in \cite{mybrief2onBLInputs2019}, where the functions \emph{Initialize}
(described in \textbf{Algorithm 2})
and \emph{AddInputs}  (described in \textbf{Algorithm 3}) are utilized.

\begin{algorithm}
\caption{:~\bf The Low-Memory Implementation of the Square-Root BLS Algorithm on Added Inputs:  Computation of Output Weights and  Increment of  New Inputs}
\begin{algorithmic}[1]
\Require    Inputs $\mathbf{X}_{{\ddot  l}}$ with labels $\mathbf{Y}_{{\ddot  l}}$, added inputs with labels;
\Ensure Output weights: the ridge solution $\mathbf{\tilde W}$;
\State  Get  (${{\mathbf{Z}}^{n}}$, ${{\mathbf{H}}^{m}}$, ${{\mathbf{W}}_{{{e}}}^n }$, ${{\mathbf{\beta }}_{{{e}}}^n} $, ${{\mathbf{W}}_{{{h}}}^m }$, ${{\mathbf{\beta }}_{{{h}}}^m}$)=
\emph{Initialize} ($\mathbf{X}_{{\ddot  l}}$);
\State   Set $\mathbf{A}_{{\ddot  l}}=\left[ {{\mathbf{Z}}^{n}}|{{\mathbf{H}}^{m}} \right]$;
\State   Utilize $\mathbf{A}_{{\ddot  l}}$ and $\mathbf{Y}_{{\ddot  l}}$  to compute ${{\mathbf{F}}_{\ddot b}}$ and ${\mathbf{\tilde W}}_{\ddot b}$  by  (\ref{FbFbLamda30929dks23cxd}) and (\ref{Wb2Ab0239ald2320akc}), respectively,
 and then obtain  ${\mathbf{F}}_{\ddot  l}  $ and $ {\mathbf{\tilde W}}_{\ddot  l}   $ by computing (\ref{LowMemChol3209dzsAll}) iteratively  for $i=2b,3b,\cdots,(l/b)b$;
\While{\emph{New inputs $\mathbf{\bar X}_{{\ddot  p}}$ and labels $\mathbf{\bar Y}_{{\ddot  p}}$ are added}}
\State  Get   (${{\mathbf{\bar Z}}_{{\ddot  p}}^{n}}$, ${{\mathbf{\bar H}}_{{\ddot  p}}^{m}}$)=\emph{AddInputs}($\mathbf{\bar X}_{{\ddot  p}}$, ${{\mathbf{W}}_{{{e}}}^n }$, ${{\mathbf{\beta }}_{{{e}}}^n} $, ${{\mathbf{W}}_{{{h}}}^m }$, ${{\mathbf{\beta }}_{{{h}}}^m}$);
\State   Set $\mathbf{\bar A}_{{{\ddot  p}}}=\left[ {{\mathbf{\bar Z}}_{{\ddot  p}}^{n}}|{{\mathbf{\bar H}}_{{\ddot  p}}^{m}} \right]$;
\State   Utilize $\mathbf{\bar A}_{{{\ddot  p}}}$ and $\mathbf{\bar Y}_{{\ddot  p}}$  to update
${{\mathbf{F}}_{\ddot l}}$  and  ${\mathbf{\tilde W}}_{\ddot l}$
into  ${{\mathbf{F}}_{\ddot l+\ddot p}}$  and  ${\mathbf{\tilde W}}_{\ddot l+\ddot p}$, respectively, by
computing (\ref{LowMemChol3209dzsAll}) iteratively for $i=l+b,l+2b,\cdots,l+(p/b)b$;
\State Set  ${{\bf{A}}_{\ddot l + \ddot p}} = {\left[ {\begin{array}{*{20}{c}}
{{\bf{A}}_{\ddot l}^T}&{{\bf{\bar A}}_{\ddot p}^T}
\end{array}} \right]^T}$  and  $l=l+p$;
\EndWhile
\State Set output weights $\mathbf{\tilde W}={\mathbf{\tilde W}}_{\ddot  l}   $;
\end{algorithmic}
\end{algorithm}

\subsection{The Proposed Low-Memory Implementation of the Square-Root BLS Algorithm on Added Nodes}

To process a batch of $b$ nodes
 in each recursion,
 we can divide the expanded input matrix $\mathbf{A}_{{{k}}}$ with $k$ nodes
 into $k/b$ batches and  each batch includes $b$ nodes, i.e., $k=(k/b)b$. Accordingly,  let us write
 \begin{equation}\label{Al2AlbAb1b20923ds23ds}
{{\bf{A}}_k}  = {\left[ {\begin{array}{*{20}{c}}
{{\bf{A}}_{b_b}}&{{\bf{A}}_{b_{2b}}}& \cdots & {{\bf{A}}_{b_{j}}}  & \cdots &{{\bf{A}}_{b_{k}}}
\end{array}} \right]},
 \end{equation}
 where ${\bf{A}}_{b_j}$ ($j=b,2b,\cdots,k$) denotes the $b$ columns of $\mathbf{A}_{{k}}$ from column $j-b+1$ to column $j$.
Then  we can
process ${{\bf{A}}_{{b_{j}}}}$ to
 update ${{\mathbf{F}}_{j-b}}$ and $\mathbf{\tilde W}_{j-b}$ into
 ${{\mathbf{F}}_{j}}$  and  ${\mathbf{\tilde W}}_{j}$, respectively,
 by simply setting $k=j-b$ and  $ q = b$   in
  (\ref{R_def_perhaps_No_R1}),
  (\ref{EFdefine13214ZF13ab}),
(\ref{ZF_def_L_2_items3ab}),
 (\ref{L_big_BLK_def1})
 and  (\ref{W2AMHEWNH3495})
to obtain
 \begin{equation}\label{R_defibib3209ds2Apr26}{{\bf{R}}_{j}}=\left[ \begin{matrix}
   {{\bf{R}}_{j-b}} & {\mathbf{\tilde R}}_{j-b,b}  \\
  {\mathbf{\tilde R}}_{j-b,b}^{T} & {{\bf{ R}}_{b_{j}}}  \\
\end{matrix} \right],
\end{equation}
\begin{subnumcases}{\label{EFdefine13214ZF13abBatch19sd}}
{\mathbf{\tilde R}}_{j-b,b} ={{\bf{A}}_{j-b}^T}\mathbf{A}_{b_{j}} &  \label{EFdefine13214ZF13aBatch19sd}\\
{{\bf{R}}_{b_{j}}} =\mathbf{A}_{b_{j}}^{T}\mathbf{A}_{b_{j}}+\lambda \mathbf{I}, & \label{EFdefine13214ZF13bBatch19sd}
\end{subnumcases}
  \begin{subnumcases}{\label{ZF_def_L_2_items3ab4batch}}
 {{\bf{ F}}_{b_{j}}} {{\bf{ F}}_{b_{j}}^{T}}  =\left({{\bf{ R}}_{b_{j}}}- {\mathbf{\tilde R}}_{j-b,b}^{T} {{\bf{F}}_{j-b}} {{\bf{F}}_{j-b}^{T}}{\mathbf{\tilde R}}_{j-b,b}\right)^{-1}&  \label{FbFbT032943aApr26}\\
 {{\bf{\tilde F}}_{j-b,b}}=-{{\bf{F}}_{j-b}}{{\bf{F}}_{j-b}^{T}}{\mathbf{\tilde R}}_{j-b,b}{{{\bf{ F}}_{b_{j}}}}, & \label{ZF_def_L_2_items3b4batch}
\end{subnumcases}
  \begin{equation}\label{LbigBLKibib3209ds3Apr26}{{\bf{F}}_{j}}=\left[ \begin{matrix}
   {{\bf{F}}_{j-b}} & {{\bf{\tilde F}}_{j-b,b}}  \\
   \mathbf{0} & {{\bf{ F}}_{b_{j}}}  \\
\end{matrix} \right]
\end{equation}
and
\begin{multline}\label{W2AMHEWNH3495bBatch320s4batch}{ {\bf{\tilde W}}_{j} }= \\ \left[ \begin{matrix}
   {{\bf{\tilde W}}_{j-b}}+{{\bf{\tilde F}}_{j-b,b}}{{\bf{ F}}_{b_{j}}^{T}}\left( \mathbf{ A}_{b_{j}}^{T}\mathbf{Y}-{\mathbf{\tilde R}}_{j-b,b}^{T}{{\bf{\tilde W}}_{j-b}} \right)  \\
   {{\bf{ F}}_{b_{j}}} {{\bf{ F}}_{b_{j}}^{T}}\left( \mathbf{ A}_{b_{j}}^{T}\mathbf{Y}-{\mathbf{\tilde R}}_{j-b,b}^{T}{{\bf{\tilde W}}_{j-b}} \right)  \\
\end{matrix} \right],
\end{multline}
 respectively,
 where
 ${{\bf{ F}}_{b_{j}}}$ is the $b \times b$ sub-matrix
 from the $(j-b+1)^{th}$ row and column to the $j^{th}$ row and column in ${{\bf{F}}_{j}}$.

Let us compute the initial ${{\bf{F}}_{{ b}}}$ and ${ {\bf{\tilde W}}_{b} }$
 by
 (\ref{L_m_def12431before}) and
(\ref{W2AMHEWNH3495initial}), respectively, where  $k$ is set to $b$.
Then we can
 compute  ${{\bf{F}}_{{j}}}$ and ${{\bf{\tilde W}}_{j}}$
from ${{\bf{F}}_{{j- b}}}$ and ${ {\bf{\tilde W}}_{j- b} }$  iteratively
by (\ref{EFdefine13214ZF13abBatch19sd}), (\ref{ZF_def_L_2_items3ab4batch}),
(\ref{LbigBLKibib3209ds3Apr26})
and
(\ref{W2AMHEWNH3495bBatch320s4batch}),
for $j=2b,3b,\cdots, (k/b)b$  to obtain ${{\bf{F}}_k}$ and ${{\bf{\tilde W}}_{k}}$ from ${{\bf{F}}_{{ b}}}$ and ${ {\bf{\tilde W}}_{b} }$,
and for~\footnote{Here ${{\bf{ A}}_{b_{j}}}$ and ${{\bf{A}}_{j-b}}$ are in ${{\bf{A}}_{k+q}}$.
}   $j=k+b,k+2b,\cdots, k+p$ to obtain ${{\bf{F}}_{k+p}}$ and ${{\bf{\tilde W}}_{k+p}}$ from  ${{\bf{F}}_k}$ and ${{\bf{\tilde W}}_{k}}$.
The above described algorithm can be summarized in the following \textbf{Algorithm 5},
which  gives the low-memory implementation of the square-root BLS algorithm on added nodes proposed in \cite{mybrief1onBLNodes2019}.


\begin{algorithm}
\caption{:~\bf The Low-Memory Implementation of the Square-Root BLS Algorithm on Added Nodes:  Computation of Output Weights and  Increment of  Nodes}
\begin{algorithmic}[1]
\Require   training sample $\mathbf{X}$ and the corresponding label $\mathbf{Y}$;
\Ensure Output weights: the ridge solution $\mathbf{\tilde W}$;
\State  Compute  (${{\mathbf{Z}}^{n}}$, ${{\mathbf{H}}^{m}}$, ${{\mathbf{W}}_{{{e}}}^n }$, ${{\mathbf{\beta }}_{{{e}}}^n} $, ${{\mathbf{W}}_{{{h}}}^m }$, ${{\mathbf{\beta }}_{{{h}}}^m}$)=
\emph{Initialize} ($\mathbf{X}$);
\State   Set $\mathbf{A}^{n,m}_k=\left[ {{\mathbf{Z}}^{n}}|{{\mathbf{H}}^{m}} \right]$;
\State  Utilize $\mathbf{A}^{n,m}_k$ and $\mathbf{Y}$ to compute the initial ${{\bf{F}}_{{ b}}}$ and ${ {\bf{\tilde W}}_{b} }$
 by (\ref{L_m_def12431before}) and
(\ref{W2AMHEWNH3495initial}), respectively, and then obtain ${{\bf{F}}_k^{n,m}}$ and ${{\bf{\tilde W}}_{k}^{n,m}}$ by computing
(\ref{EFdefine13214ZF13abBatch19sd}),
(\ref{ZF_def_L_2_items3ab4batch}),
(\ref{LbigBLKibib3209ds3Apr26})
and
(\ref{W2AMHEWNH3495bBatch320s4batch}) iteratively for $j=2b,3b,\cdots, k$;
\While{\emph{The target training error  is not reached}}
\If{\emph{only enhancement nodes are added}}
\State  Random   ${{\mathbf{W}}_{{{h}_{m+1}}}}$ and
    ${{\bm{\beta }}_{{{h}_{m+1}}}}$;
\State  Compute $\mathbf{\bar A}_{q}={{\mathbf{H}}_{m+1}}=\xi ({{\mathbf{Z}}^{n}}{{\mathbf{W}}_{{{h}_{m+1}}}}+{{\bm{\beta }}_{{{h}_{m+1}}}})$;
\State Obtain ${{\mathbf{F}}_{k+p}^{n,m+1}}$ and ${\mathbf{\tilde W}}_{k+p}^{n,m+1}$  by  computing (\ref{EFdefine13214ZF13abBatch19sd}), \\
\quad \quad  (\ref{ZF_def_L_2_items3ab4batch}),  (\ref{LbigBLKibib3209ds3Apr26}) and
(\ref{W2AMHEWNH3495bBatch320s4batch})  for $j=k+b,k+2b,\cdots,k+p$;
\State Set  ${\mathbf{A}}_{k+q}^{n,m+1}=\left[ {{\mathbf{A}}_k^{n,m}}|\mathbf{\bar A}_{q}\right]$;
\State Set  $m=m+1$ and $k=k+p$;
\Else[\emph{feature nodes are added}]
\State  Fine-tuned random ${{\mathbf{W}}_{{{e}_{n+1}}}}$ and  ${{\bm{\beta }}_{{{e}_{n+1}}}}$;
\State Compute ${{\mathbf{Z}}_{n+1}}=\phi (\mathbf{X}{{\mathbf{W}}_{{{e}_{n+1}}}}+{{\bm{\beta }}_{{{e}_{n+1}}}})$;
\State Set random ${{\mathbf{W}}_{{ex}_{i}}}$ and ${{\bm{\beta }}_{{ex}_{i}}}$ for \\
\quad \quad  $i=1,2,\cdots,m$ and compute
       $\scriptstyle {{\mathbf{H}}_{{ex}_{n+1}}}=$ \\
       \quad \quad   $\scriptstyle \left[\xi ({{\mathbf{Z}}_{n+1}}{{\mathbf{W}}_{{ex}_{1}}}+{{\bm{\beta }}_{{ex}_{1}}}),\cdots ,\xi ({{\mathbf{Z}}_{n+1}}{{\mathbf{W}}_{{ex}_{m}}}+{{\bm{\beta }}_{{ex}_{m}}})\right]$;
\State Set $\mathbf{\bar A}_{q}=\left[  {{\mathbf{Z}}_{n+1}}|{{\mathbf{H}}_{e{{x}_{n+1}}}} \right] $;
 \State Obtain ${{\mathbf{F}}_{k+p}^{n+1,m}}$ and ${\mathbf{\tilde W}}_{k+p}^{n+1,m}$  by  computing (\ref{EFdefine13214ZF13abBatch19sd}),   \\
\quad \quad  (\ref{ZF_def_L_2_items3ab4batch}),  (\ref{LbigBLKibib3209ds3Apr26}) and
(\ref{W2AMHEWNH3495bBatch320s4batch})  for $j=k+b,k+2b,\cdots,k+p$;
\State Set  ${\mathbf{A}}_{k+q}^{n+1,m} = \left[ {{\mathbf{A}}_k^{n,m}}|\mathbf{\bar A}_{q}\right]$;
\State  Set $n=n+1$ and $k=k+p$;
\EndIf
\EndWhile
\State Set output weights $\mathbf{\tilde W}={\mathbf{\tilde W}}_{k}^{n,m}$;
\end{algorithmic}
\end{algorithm}

%
%
%
%

\section{The Proposed Full Low-Memory Square-Root  BLS Implementation Based on Inverse Cholesky Factorizations with a Batch Size $b$}

In the last section, we have developed the low-memory implementations for
 the recursive and square-root BLS algorithms on added inputs
  proposed in  \cite{mybrief2onBLInputs2019} and  the square-root BLS algorithm on added nodes
  proposed in  \cite{mybrief1onBLNodes2019}.
  It is well known that in the processor units with limited
precision,
   the recursive updates of
    the inverse matrix
may introduce numerical instabilities
   after a large number of iterations~\cite{TransSP2003Blast},
 as will be verified by the numerical experiments in the next section.
 On the other hand,
 when cooperating with the low-memory implementation of the square-root BLS algorithm on added nodes~\cite{mybrief1onBLNodes2019},
 the  low-memory implementation of  the recursive BLS algorithm on added inputs needs the extra computational load to decompose the
 inverse matrix  into the Cholesky factor.
 Accordingly, in this section we only discuss the  low-memory implementations of the square-root BLS algorithms on added inputs and nodes,
 to propose the full low-memory implementation of the square-root BLS algorithm.

 \subsection{Inverse Cholesky Factorizations with a Batch Size $b$}

 To obtain~\footnote{Here the matrices include both the dotted subscript and the subscript, which
 indicate the numbers of  training samples and nodes, respectively.}  ${{\bf{F}}_{{\ddot  l},k}^{n,m}}$ and ${{\bf{\tilde W}}_{{\ddot  l},k}^{n,m}}$,
 it requires $l/b-1$ iterations to compute  (\ref{LowMemChol3209dzsAll}) iteratively  for $i=2b,3b,\cdots,(l/b)b$ in \textbf{Algorithm 4},
 and it requires $k/b-1$ iterations to compute
 (\ref{ZF_def_L_2_items3ab4batch}),
(\ref{LbigBLKibib3209ds3Apr26})
and
(\ref{W2AMHEWNH3495bBatch320s4batch}) iteratively for $j=2b,3b,\cdots, k$ in \textbf{Algorithm 5}.
Usually there are always  more  training samples than nodes in BLS~\cite{27_ref_BL_trans_paper, BL_trans_paper,mybrief2onBLInputs2019},
i.e., $l>k$. Then it can be seen that
 \textbf{Algorithm 4} requires
 more iterations to compute ${{\bf{F}}_{{\ddot  l},k}^{n,m}}$ and ${{\bf{\tilde W}}_{{\ddot  l},k}^{n,m}}$
 than  \textbf{Algorithm 5}.   
 Furthermore, it is required to update all the $\frac{1}{2}k^2$ entries of ${{\mathbf{F}}_{\ddot i- \ddot b}} \in {\Re ^{k \times k}}$  to compute  ${{\mathbf{F}}_{\ddot i }} \in {\Re ^{k \times k}}$
 from ${{\mathbf{F}}_{\ddot i- \ddot b}}$ by  (\ref{LowMemChol3209dzs4}) in \textbf{Algorithm 4},
 while it is only required to  add  a $j \times b$ sub-matrix
 to the right
 side of ${{\bf{F}}_{j-b}} \in {\Re ^{(j-b) \times (j-b)}} $ to update
 ${{\bf{F}}_{j-b}}$ into ${{\bf{F}}_{j}} \in {\Re ^{j \times j}} $ by
 (\ref{LbigBLKibib3209ds3Apr26}) for $j=2b,3b,\cdots, k$ in \textbf{Algorithm 5}.
Accordingly, it can be concluded that
\textbf{Algorithm 4} requires
 more computation complexities to compute ${{\bf{F}}_{{\ddot  l},k}^{n,m}}$ and ${{\bf{\tilde W}}_{{\ddot  l},k}^{n,m}}$
 than  \textbf{Algorithm 5}.


Since \textbf{Algorithm 5} is more efficient than {\textbf{Algorithm 4}},
  we utilize  (\ref{L_m_def12431before})  (with $k=b$),
 (\ref{R_defibib3209ds2Apr26}),
 (\ref{ZF_def_L_2_items3ab4batch})
 and
(\ref{LbigBLKibib3209ds3Apr26}) that are relevant to  the computation of
${{\bf{F}}_{k}}$  (i.e.,  ${{\bf{F}}_{{\ddot  l},k}^{n,m}}$)
in \textbf{Algorithm 5},
 to
  develop
 the function
\begin{equation}\label{Fk2InvChol1basedOnR203392a12}
{{\bf{F}}_k}=\emph{InvChol}(\mathbf{R}_{k}, b)
 \end{equation}
 described by \textbf{Algorithm 6}.
 It can be seen that
\textbf{Algorithm 6}
obtains
${{\bf{F}}_k}=\emph{InvChol}(\mathbf{R}_{k}, b)$ satisfying
 (\ref{L_m_def12431before}) (which is  ${{\bf{F}}_k} {{\bf{F}}_k^T}=\mathbf{R}_{k}^{-1}=({{\bf{A}}_k^T}{{\bf{A}}_k}+ \lambda {\bf{I}})^{-1}$), i.e.,
 \textbf{Algorithm 6} computes the inverse Cholesky factor of a Hermitian Matrix $\mathbf{R}_{k}$.
 It can be seen that in \textbf{Algorithm 6} with a batch size of $b$, only the inverse Cholesky factorizations of
 $b \times b$ matrices are required, and the inversions of large matrices are not required.

 \begin{algorithm}
\caption{The Proposed Inverse Cholesky Factorization of a Hermitian Matrix $\mathbf{R}_{k}$ with a Batch Size of $b$}\label{euclid}
\begin{algorithmic}[0]
\Function{${InvChol}$}{$\mathbf{R}_{k}$, $b$}
\State Compute the initial inverse Cholesky factor ${{\bf{F}}_{b}}$ by \\ \quad   ${{\bf{F}}_{b}}{{\bf{F}}_{b}^{T}} ={{\bf{R}}_{b}^{-1} }$;
\For{$j=2b,3b,\cdots, (k/b)b$}
\State Obtain
${{\bf{ R}}_{b_{j}}}$
 and  ${\mathbf{\tilde R}}_{j-b,b}$
in  ${{\bf{R}}_{j}}$ by \\
\quad \quad \,  ${{\bf{R}}_{j}}=\left[ \begin{matrix}
   {{\bf{R}}_{j-b}} & {\mathbf{\tilde R}}_{j-b,b}  \\
  {\mathbf{\tilde R}}_{j-b,b}^{T} & {{\bf{ R}}_{b_{j}}}  \\
\end{matrix} \right]$;
\State Compute the inverse Cholesky factor
 ${{\bf{ F}}_{b_{j}}}$  by \\ \quad \quad \, ${{\bf{ F}}_{b_{j}}} {{\bf{ F}}_{b_{j}}^{T}}  =\left({{\bf{ R}}_{b_{j}}}- {\mathbf{\tilde R}}_{j-b,b}^{T} {{\bf{F}}_{j-b}} {{\bf{F}}_{j-b}^{T}}{\mathbf{\tilde R}}_{j-b,b}\right)^{-1}$;
\State Compute ${{\bf{\tilde F}}_{j-b,b}}=-{{\bf{F}}_{j-b}}{{\bf{F}}_{j-b}^{T}}{\mathbf{\tilde R}}_{j-b,b}{{{\bf{ F}}_{b_{j}}}}$;
\State Obtain ${{\bf{F}}_{j}}=\left[ \begin{matrix}
   {{\bf{F}}_{j-b}} & {{\bf{\tilde F}}_{j-b,b}}  \\
   \mathbf{0} & {{\bf{ F}}_{b_{j}}}  \\
\end{matrix} \right]$;
\EndFor
\State \textbf{return} ${{\bf{F}}_{(k/b)b}}={{\bf{F}}_{k}}$;
\EndFunction
\end{algorithmic}
\end{algorithm}

 Since we are developing low-memory implementations in this paper,
let us
  save the memories for storing ${{\bf{R}}_k}$ in \textbf{Algorithm 6}, by substituting (\ref{EFdefine13214ZF13abBatch19sd}) into (\ref{ZF_def_L_2_items3ab4batch})
to obtain
\begin{subnumcases}{\label{ZF_def_L_2_items3abNoRBatch112}}
{{{\bf{ F}}_{b_{j}}}}{{\bf{ F}}_{b_{j}}^{T}}=\left( \begin{array}{l}
\mathbf{ A}_{b_{j}}^{T}\mathbf{ A}_{b_{j}}+\lambda \mathbf{I}- {\mathbf{ A}_{b_{j}}^T} \times \\
 {{\bf{A}}_{j-b}}
{{\bf{F}}_{j-b}}{{\bf{F}}_{j-b}^{T}}{{\bf{A}}_{j-b}^T}\mathbf{ A}_{b_{j}}
\end{array} \right)^{-1} &  \label{ZF_def_L_2_items3aNoRBatch112}\\
 {{\bf{\tilde F}}_{j-b,b}}=-{{\bf{F}}_{j-b}}{{\bf{F}}_{j-b}^{T}}{{\bf{A}}_{j-b}^T}\mathbf{ A}_{b_{j}}{{{\bf{F}}_{b_{j}}}}. & \label{ZF_def_L_2_items3bNoRBatch112}
\end{subnumcases}
Then (\ref{ZF_def_L_2_items3abNoRBatch112}),
  (\ref{L_m_def12431before})  (with $k=b$)
   and
(\ref{LbigBLKibib3209ds3Apr26}) are utilized to develop
 the function
 \begin{equation}\label{Fk2InvChol120ks2s23}
{{\bf{F}}_k}={\Phi}_{chol}(\mathbf{A}_{k},\lambda, b)
 \end{equation}
described in \textbf{Algorithm 7}, which utilizes $\mathbf{A}_{k}$ to compute  the upper-triangular inverse Cholesky factor of
the $k \times k$ Hermitian matrix
${{\bf{A}}_k^T}{{\bf{A}}_k}+ \lambda {\bf{I}}$.

 \begin{algorithm}
\caption{The Proposed Inverse Cholesky Factorization of  ${{\bf{A}}_k^T}{{\bf{A}}_k}+ \lambda {\bf{I}}$ with a Batch Size $b$}\label{euclid}
\begin{algorithmic}[0]
\Function{${\Phi}_{chol}$}{$\mathbf{A}_{k}$, $\lambda$, $b$}
\State Compute the initial inverse Cholesky factor ${{\bf{F}}_{b}}$ by \\ \quad   ${{\bf{F}}_{b}}{{\bf{F}}_{b}^{T}} =
({{\bf{A}}_b^T}{{\bf{A}}_b}+ \lambda {\bf{I}})^{-1}$;
\For{$j=2b,3b,\cdots, (k/b)b$}
\State Compute the inverse Cholesky factor
 ${{\bf{ F}}_{b_{j}}}$  by \\ \quad \quad \, ${{{\bf{ F}}_{b_{j}}}}{{\bf{ F}}_{b_{j}}^{T}}=\left( \begin{array}{l}
\mathbf{ A}_{b_{j}}^{T}\mathbf{ A}_{b_{j}}+\lambda \mathbf{I}- {\mathbf{ A}_{b_{j}}^T} \times \\
 {{\bf{A}}_{j-b}}
{{\bf{F}}_{j-b}}{{\bf{F}}_{j-b}^{T}}{{\bf{A}}_{j-b}^T}\mathbf{ A}_{b_{j}}
\end{array} \right)^{-1}$;
\State Compute ${{\bf{\tilde F}}_{j-b,b}}=-{{\bf{F}}_{j-b}}{{\bf{F}}_{j-b}^{T}}{{\bf{A}}_{j-b}^T}\mathbf{ A}_{b_{j}}{{{\bf{F}}_{b_{j}}}}$;
\State Obtain ${{\bf{F}}_{j}}=\left[ \begin{matrix}
   {{\bf{F}}_{j-b}} & {{\bf{\tilde F}}_{j-b,b}}  \\
   \mathbf{0} & {{\bf{ F}}_{b_{j}}}  \\
\end{matrix} \right]$;
\EndFor
\State \textbf{return} ${{\bf{F}}_{(k/b)b}}={{\bf{F}}_{k}}$;
\EndFunction
\end{algorithmic}
\end{algorithm}






\subsection{The Proposed Full Low-Memory Square-Root  BLS Implementation}

 To avoid the out-of-memory problems caused by the inversions of large matrices,
 we can
 utilize (\ref{Fk2InvChol1basedOnR203392a12}) or (\ref{Fk2InvChol120ks2s23}) to
 compute the inverse Cholesky factorizations in  the square-root BLS algorithm on added nodes~\cite{mybrief1onBLNodes2019}
  and
that   on added inputs~\cite{mybrief2onBLInputs2019}, as will be introduced in what follows.

In the square-root BLS algorithm  on added nodes~\cite{mybrief1onBLNodes2019}, we  utilize (\ref{Fk2InvChol120ks2s23})
to  compute
${{\bf{F}}_k}$ from
 $\mathbf{A}_{k}$ directly,
 and then we  can save the memories for storing ${{\bf{R}}_k}$.
We can also save the memories for storing ${{\bf{R}}_{k+q}}$,
by substituting
  (\ref{EFdefine13214ZF13ab}) into
(\ref{ZF_def_L_2_items3ab}) to obtain
\begin{subnumcases}{\label{ZF_def_L_2_items3abNoR3290as}}
{{{\bf{\bar F}}_q}}{{\bf{\bar F}}_q^{T}}=\left(\mathbf{\bar A}_{q}^{T}\mathbf{\bar A}_{q}+\lambda \mathbf{I}-
 {\mathbf{\bar A}_{q}^T} {{\bf{A}}_k}
{{\bf{F}}_k}{{\bf{F}}_k^{T}}{{\bf{A}}_k^T}\mathbf{\bar A}_{q}\right)^{-1} &  \label{ZF_def_L_2_items3aNoR3290as}\\
 {{\bf{\tilde F}}_{k,q}}=-{{\bf{F}}_k}{{\bf{F}}_k^{T}}{{\bf{A}}_k^T}\mathbf{\bar A}_{q}{{{\bf{\bar F}}_q}}. & \label{ZF_def_L_2_items3bNoR3290as}
\end{subnumcases}
To obtain the upper-triangular inverse Cholesky factor ${{{\bf{\bar F}}_q}}$ satisfying (\ref{ZF_def_L_2_items3aNoR3290as}),
we can utilize (\ref{Fk2InvChol1basedOnR203392a12}) to compute
\begin{equation}\label{Fk2InvChol13kd2NoR02393kd1}
{{{\bf{\bar F}}_q}}=\emph{InvChol}\left(\left( \begin{array}{l}
{\bf{\bar A}}_q^T{{{\bf{\bar A}}}_q} + \lambda {\bf{I}} - \\
{\bf{\bar A}}_q^T{{\bf{A}}_k}{{\bf{F}}_k}{\bf{F}}_k^T{\bf{A}}_k^T{{{\bf{\bar A}}}_q}
\end{array} \right), b\right).
 \end{equation}





In  the square-root BLS algorithm on added inputs~\cite{mybrief2onBLInputs2019},
the initial  ${\mathbf{F}}_{\ddot  l,k}$ satisfying (\ref{Q2PiPiT9686954})  can also
be computed from $\mathbf{A}_{\ddot  l, k}$ directly by
 (\ref{Fk2InvChol120ks2s23}).
When  $p<k$,  we utilize (\ref{Fk2InvChol120ks2s23}) to
compute
\begin{equation}\label{Fk2InvChol13kd2Use1}
\pmb{\Gamma}={\Phi}_{chol}(\mathbf{S},1, b)
 \end{equation}
 satisfying
 $\pmb{\Gamma} {{\pmb{\Gamma}}^T} = {{(\mathbf{I}+{{\mathbf{S}}^{T}}\mathbf{S})}^{-1}} $, which is then substituted  into
 (\ref{VWpSmallThaNk32230}) to obtain
 \begin{subnumcases}{\label{VWpSmallThaNk32230LowMem}}
{\mathbf{V}} {\mathbf{V}}^{T}= \mathbf{I}- \mathbf{S} \pmb{\Gamma} {{\pmb{\Gamma}}^T} {{\mathbf{S}}^{T}} &  \label{Lwave2IKK40425bLowMem}\\
{\mathbf{\tilde W}}_{\ddot  l    + \ddot  p   } = {{{\bf{\tilde W}}}_{\ddot  l   }} + {\mathbf{F}}_{\ddot  l   } \mathbf{S} \pmb{\Gamma} {{\pmb{\Gamma}}^T} ({{{\mathbf{\bar Y}}_{ \ddot  p   }} - \mathbf{\bar A}_{ \ddot  p   }^{}{{{\bf{\tilde W}}}_{\ddot  l   }}}).   &  \label{WfromF3209daLowMem}
\end{subnumcases}
When $p \ge k$,
 we  utilize (\ref{Fk2InvChol120ks2s23}) to
compute
\begin{equation}\label{Fk2InvChol13kd2Use1VBigp32s}
{\mathbf{V}}={\Phi}_{chol}({{\mathbf{S}}^{T}},1, b)
 \end{equation}
 satisfying (\ref{Lwave2IKK40425a}) (i.e., ${\mathbf{V}} {\mathbf{V}}^{T} ={{(\mathbf{I}+\mathbf{S} {{\mathbf{S}}^{T}})}^{-1}}$).

The above-described  low-memory implementations of the square-root BLS algorithm on added nodes~\cite{mybrief1onBLNodes2019}
and
that   on added inputs~\cite{mybrief2onBLInputs2019} are summarized in the following \textbf{Algorithm 8},
which are based on (\ref{Fk2InvChol1basedOnR203392a12}) and (\ref{Fk2InvChol120ks2s23}),
  the inverse Cholesky factorizations with a batch size $b$.
It can be seen that \textbf{Algorithm 8} gives the  full low-memory implementation of the square-root BLS Algorithm,
which includes the increment of  feature nodes, enhancement nodes and
new inputs.


\begin{algorithm}
\caption{:~\bf The Proposed Full Low-Memory Implementation of the Square-Root BLS Algorithm:  Computation of Output Weights and  Increment of  Feature Nodes, Enhancement Nodes and
New Inputs}
\begin{algorithmic}[1]
\Require Inputs $\mathbf{X}_{{\ddot  l}}$ with labels $\mathbf{Y}_{{\ddot  l}}$, added inputs with labels;
\Ensure Output weights:   the ridge solution $\mathbf{\tilde W}$;
\State  Get  (${{\mathbf{Z}}^{n}}$, ${{\mathbf{H}}^{m}}$, ${{\mathbf{W}}_{{{e}}}^n }$, ${{\mathbf{\beta }}_{{{e}}}^n} $, ${{\mathbf{W}}_{{{h}}}^m }$, ${{\mathbf{\beta }}_{{{h}}}^m}$)=
\emph{Initialize} ($\mathbf{X}_{{\ddot  l}}$);
\State   Set $\mathbf{A}^{n,m}_{{\ddot  l},k}=\left[ {{\mathbf{Z}}^{n}}|{{\mathbf{H}}^{m}} \right]$;
\State   Compute $\mathbf{F}^{n,m}_{{\ddot  l},k}={\Phi}_{chol}(\mathbf{A}^{n,m}_{{\ddot  l},k},\lambda, b)$;
\State   Compute   $\mathbf{\tilde W}^{n,m}_{{\ddot  l},k}= \mathbf{F}^{n,m}_{{\ddot  l},k}{(\mathbf{F}^{n,m}_{{\ddot  l},k})^{T}} \mathbf{A}^{n,m}_{{\ddot  l},k}  \mathbf{Y}_{{\ddot  l}}$;
\While{\emph{The target training error  is not reached}}
\If {Only enhancement nodes are added}
\State  Random   ${{\mathbf{W}}_{{{h}_{m+1}}}}$ and
    ${{\bm{\beta }}_{{{h}_{m+1}}}}$;
\State  Compute $\mathbf{\bar A}_{q}={{\mathbf{H}}_{m+1}}=\xi ({{\mathbf{Z}}^{n}}{{\mathbf{W}}_{{{h}_{m+1}}}}+{{\bm{\beta }}_{{{h}_{m+1}}}})$;
\State Utilize ${{\mathbf{A}}_{{\ddot  l},k}^{n,m}}$, $\mathbf{\bar A}_{q}$ and $\mathbf{F}^{n,m}_{{\ddot  l},k}$ to update $\mathbf{F}^{n,m}_{{\ddot  l},k}$ into
\\ \quad \quad \quad ${{\mathbf{F}}_{{\ddot  l},{k+p}}^{n,m+1}}$   by
(\ref{Fk2InvChol13kd2NoR02393kd1}),
(\ref{ZF_def_L_2_items3bNoR3290as})
and
(\ref{L_big_BLK_def1});
\State  Update $\mathbf{\tilde W}^{n,m}_{{\ddot  l},k}$ into  ${\mathbf{\tilde W}}_{{\ddot  l},{k+p}}^{n,m+1}$  by (\ref{W2AMHEWNH3495});
\State Set ${\mathbf{A}}_{{\ddot  l},k+q}^{n,m+1}=\left[ {{\mathbf{A}}_{{\ddot  l},k}^{n,m}}|\mathbf{\bar A}_{q}\right]$;
\State Set $m=m+1$ and $k=k+p$;
\ElsIf {Feature nodes are added}
\State  Fine-tuned random ${{\mathbf{W}}_{{{e}_{n+1}}}}$ and  ${{\bm{\beta }}_{{{e}_{n+1}}}}$;
\State Compute ${{\mathbf{Z}}_{n+1}}=\phi (\mathbf{X}_{{\ddot  l}}{{\mathbf{W}}_{{{e}_{n+1}}}}+{{\bm{\beta }}_{{{e}_{n+1}}}})$;
 \State Set random ${{\mathbf{W}}_{{ex}_{i}}}$ and ${{\bm{\beta }}_{{ex}_{i}}}$ for \\
\quad \quad \quad  $i=1,2,\cdots,m$ and compute
       $\scriptstyle {{\mathbf{H}}_{{ex}_{n+1}}}=$ \\
       \quad \quad \quad  $\scriptstyle \left[\xi ({{\mathbf{Z}}_{n+1}}{{\mathbf{W}}_{{ex}_{1}}}+{{\bm{\beta }}_{{ex}_{1}}}),\cdots ,\xi ({{\mathbf{Z}}_{n+1}}{{\mathbf{W}}_{{ex}_{m}}}+{{\bm{\beta }}_{{ex}_{m}}})\right]$;
\State Set $\mathbf{\bar A}_{q}=\left[  {{\mathbf{Z}}_{n+1}}|{{\mathbf{H}}_{e{{x}_{n+1}}}} \right] $;
 \State Utilize ${{\mathbf{A}}_{{\ddot  l},k}^{n,m}}$, $\mathbf{\bar A}_{q}$ and $\mathbf{F}^{n,m}_{{\ddot  l},k}$ to
 update $\mathbf{F}^{n,m}_{{\ddot  l},k}$ into
\\ \quad \quad \quad
${{\mathbf{F}}_{{\ddot  l},k+p}^{n+1,m}}$ by
(\ref{Fk2InvChol13kd2NoR02393kd1}),
(\ref{ZF_def_L_2_items3bNoR3290as})
and
(\ref{L_big_BLK_def1});
\State   Update $\mathbf{\tilde W}^{n,m}_{{\ddot  l},k}$ into
 ${\mathbf{\tilde W}}_{{\ddot  l},k+p}^{n+1,m}$  by (\ref{W2AMHEWNH3495});
\State Set  ${\mathbf{A}}_{{\ddot  l},k+q}^{n+1,m} = \left[ {{\mathbf{A}}_{{\ddot  l},k}^{n,m}}|\mathbf{\bar A}_{q}\right]$;
\State  Set $n=n+1$ and $k=k+p$;
\ElsIf {\emph{New inputs $\mathbf{\bar X}_{{\ddot  p}}$ and labels $\mathbf{\bar Y}_{{\ddot  p}}$ are added}}
\State    (${{\mathbf{\bar Z}}_{{\ddot  p}}^{n}}$, ${{\mathbf{\bar H}}_{{\ddot  p}}^{m}}$)=\emph{AddInputs}($\mathbf{\bar X}_{{\ddot  p}}$, ${{\mathbf{W}}_{{{e}}}^n }$, ${{\mathbf{\beta }}_{{{e}}}^n} $, ${{\mathbf{W}}_{{{h}}}^m }$, ${{\mathbf{\beta }}_{{{h}}}^m}$);
\State   Set $\mathbf{\bar A}_{{{\ddot  p}}}=\left[ {{\mathbf{\bar Z}}_{{\ddot  p}}^{n}}|{{\mathbf{\bar H}}_{{\ddot  p}}^{m}} \right]$;
\State Compute the intermediate result $\mathbf{S}$ by (\ref{K2AxLm94835});
\If{$p<k$}
\State Compute $\pmb{\Gamma}$ and ${\mathbf{V}}$
by
 (\ref{Fk2InvChol13kd2Use1}) and
(\ref{Lwave2IKK40425bLowMem});
\State Update ${{\mathbf{\tilde W}}_{{\ddot  l},k}^{n,m}}$ into ${\mathbf{\tilde W}}_{{\ddot  l + \ddot p},k}^{n,m}$
 by
(\ref{WfromF3209daLowMem});
\State   Update ${{\mathbf{F}}_{{\ddot  l},k}^{n,m}}$ into ${\mathbf{F}}_{{\ddot  l + \ddot p},k}^{n,m}$
 by (\ref{Lbig2LLwave59056});
\Else[$p \ge k$]
\State  Compute ${\mathbf{V}}$  by (\ref{Fk2InvChol13kd2Use1VBigp32s});
\State Update   ${{\mathbf{F}}_{{\ddot  l},k}^{n,m}}$ into ${\mathbf{F}}_{{\ddot  l + \ddot p},k}^{n,m}$
 by
(\ref{Lbig2LLwave59056});
\State  Update
 ${{\mathbf{\tilde W}}_{{\ddot  l},k}^{n,m}}$ into ${\mathbf{\tilde W}}_{{\ddot  l + \ddot p},k}^{n,m}$
 by
(\ref{WfromF3209db});
\EndIf
\State Set ${\mathbf{A}}_{{\ddot  l + \ddot p},k}^{n,m} = {\left[ {\begin{array}{*{20}{c}}
{({\mathbf{A}}_{{\ddot  l},k}^{n,m})^T}&{{\bf{\bar A}}_{\ddot p}^T}
\end{array}} \right]^T}$ and $l=l+p$;
\EndIf
\EndWhile
\State $\mathbf{\tilde W}={{\mathbf{\tilde W}}_{{\ddot  l},k}^{n,m}}$;
\end{algorithmic}
\end{algorithm}

\begin{table*}[!t]
\renewcommand{\arraystretch}{1.3}
\newcommand{\tabincell}[2]{\begin{tabular}{@{}#1@{}}#2\end{tabular}}
\caption{Snapshot Results of Testing Accuracy for Presented $4$ BLS Algorithms on Added Inputs} \label{table_example} \centering
\begin{tabular}{|c||c c c c| c c c c|}
\hline
 {\bfseries   \tabincell{c}{Number }}
      &\multicolumn{8}{c|}{{\bfseries   \tabincell{c}{ Testing  Accuracy ($\% $) }}}
         \\
   {\bfseries   \tabincell{c}{ of }}
      &\multicolumn{4}{c|}{ {\bfseries   { $\lambda=10^{-8}$}} }
       &\multicolumn{4}{c|}{{\bfseries   { $\lambda=1/128$}}}
        \\
  {\bfseries   \tabincell{c}{ Input Patterns  }}    &Exist        &Alg.3       &Alg.4           &Alg.8             &Exist        &Alg.3       &Alg.4           &Alg.8        \\
\hline
  \bfseries  15000  &    10.02   &   89.87   &   89.84   &   89.84   &   91.56   &   91.70   &  91.56   &   91.56           \\
 \bfseries  15000 $\xrightarrow[\scriptscriptstyle{9000}]{}$ 24000  &     10.02   &   95.00   &   94.95   &   94.95   &   93.81   &   95.27   &  95.04   &   95.04             \\
 \bfseries 24000 $\xrightarrow[\scriptscriptstyle{9000}]{}$ 33000   &    10.02   &   95.90   &   96.01   &   96.01   &   94.66   &   96.23   &   96.04   &   96.04           \\
 \bfseries 33000 $\xrightarrow[\scriptscriptstyle{9000}]{}$ 42000   &    10.02   &   96.00   &   96.42   &   96.42   &   95.33   &   96.49   &   96.40   &   96.40            \\
 \bfseries 42000 $\xrightarrow[\scriptscriptstyle{9000}]{}$ 51000    &     10.02   &   96.60   &   96.71   &   96.71   &   95.57   &   96.75   &   96.72   &   96.72          \\
 \bfseries 51000 $\xrightarrow[\scriptscriptstyle{9000}]{}$ 60000   &     10.02   &    8.92   &   96.86   &   96.86   &   95.89   &   96.85   &    96.87   &   96.87             \\
\hline
\end{tabular}
\end{table*}

 \begin{table*}[!t]
\renewcommand{\arraystretch}{1.3}
\newcommand{\tabincell}[2]{\begin{tabular}{@{}#1@{}}#2\end{tabular}}
\caption{Snapshot Results of Testing Accuracy for Proposed $2$ BLS Algorithms on Added Nodes} \label{table_example} \centering
\begin{tabular}{|c||c|c c | c c |}
\hline
 {\bfseries   \tabincell{c}{Number }}
      &\multicolumn{5}{c|}{{\bfseries   \tabincell{c}{  Testing  Accuracy ($\% $) }}}
         \\
   {\bfseries   \tabincell{c}{ of }}
   & {{\bfseries   { $\lambda=10^{-8}$}}}
      &\multicolumn{2}{c|}{{\bfseries   { $\lambda=10^{-7}$}}}
       &\multicolumn{2}{c|}{{\bfseries   { $\lambda=1/128$}}}
        \\
  {\bfseries   \tabincell{c}{ Total Nodes  }}    &Alg. 8       &Alg. 5        &Alg. 8            &Alg. 5        &Alg. 8            \\
\hline
  \bfseries 3060  &98.40  & 98.39   &   98.39   &   95.53   &   95.53      \\
 \bfseries 3060 $\xrightarrow[\scriptscriptstyle{2010}]{}$ 5070  &98.60  & 98.57   &   98.57   &   96.16   &   96.16     \\
 \bfseries 5070 $\xrightarrow[\scriptscriptstyle{2010}]{}$ 7080  &98.81  & 98.79   &   98.79   &   96.36   &   96.36      \\
 \bfseries 7080 $\xrightarrow[\scriptscriptstyle{2010}]{}$ 9090 &98.91  & 98.87   &   98.87   &   96.57   &   96.57     \\
 \bfseries 9090 $\xrightarrow[\scriptscriptstyle{2010}]{}$ 11100 &98.91   &  98.82   &   98.83   &   96.78   &   96.78     \\
\hline
\end{tabular}
\end{table*}

\section{Numerical Experiments}

 Numerical experiments will be carried out in this section,
to compare the presented low-memory BLS implementations.
The
presented implementations on added inputs include
the existing one  proposed
 in \cite{BLSLowMemTNNLS2020dec},
 the part for added inputs in \textbf{Algorithm 8},
 and the  low-memory implementations of
the recursive and square-root BLS algorithms in \cite{mybrief2onBLInputs2019}( i.e.,
  \textbf{Algorithm 3}  and \textbf{Algorithm 4}).
On the other hand,  the
presented
implementations on added nodes include
 the low-memory implementation of the square-root BLS algorithm in \cite{mybrief1onBLNodes2019} (i.e., \textbf{Algorithm 5})
and the part for added nodes in \textbf{Algorithm 8}.
Notice that  \textbf{Algorithm 8} is the proposed  full low-memory implementation of the square-root BLS Algorithm,
which includes the part for added inputs and that for added nodes.


\begin{table*}[!t]
\renewcommand{\arraystretch}{1.3}
\newcommand{\tabincell}[2]{\begin{tabular}{@{}#1@{}}#2\end{tabular}}
\caption{Snapshot Results of Training Time for Presented $4$ BLS Algorithms on Added Inputs and the Corresponding Time Ratios, where the  Newly Added
Inputs are Less than the Nodes of the Network.} \label{table_example} \centering
\begin{tabular}{|c||c c c c| c c c|c c c c|c c c|}
\hline
 {\bfseries   \tabincell{c}{Number }}
      &\multicolumn{7}{c|}{{\bfseries   \tabincell{c}{ Additional  Training Time }}}
       &\multicolumn{7}{c|}{{\bfseries   \tabincell{c}{Accumulative Training Time }}}
         \\
   {\bfseries   \tabincell{c}{ of }}
      &\multicolumn{4}{c|}{{\bfseries   \tabincell{c}{ Time in  Seconds }}}
      &\multicolumn{3}{c|}{{\bfseries   \tabincell{c}{  Time Ratios  to Exist}}}
       &\multicolumn{4}{c|}{{\bfseries   \tabincell{c}{ Time in  Seconds }}}
      &\multicolumn{3}{c|}{{\bfseries   \tabincell{c}{ Time Ratios to Exist}}}  \\
  {\bfseries   \tabincell{c}{ Input Patterns  }}    &Exist        &Alg.3       &Alg.4           &Alg.8         &Alg.3       &Alg.4           &Alg.8        &Exist        &Alg.3       &Alg.4           &Alg.8         &Alg.3       &Alg.4           &Alg.8           \\
\hline
  \bfseries  15000  &           213   &         194   &        5065   &         182   &          0.91   &  23.78   &          0.85   &         213   &         194   &        5065   &    182   &          0.91   &        23.78   &          0.85  \\
 \bfseries  15000 $\xrightarrow[\scriptscriptstyle{9000}]{}$ 24000  &    173   &         124   &        3055   &         197   &          0.72   &  17.66   &         1.14   &         386   &         318   &        8120   &   379   &          0.82   &        21.04   &          0.98     \\
 \bfseries 24000 $\xrightarrow[\scriptscriptstyle{9000}]{}$ 33000   &   134   &         133   &        3507   &         349   &          0.99   &  26.17   &         2.60   &         520   &         451   &       11627   &   728   &          0.87   &        22.36   &         1.40    \\
 \bfseries 33000 $\xrightarrow[\scriptscriptstyle{9000}]{}$ 42000    &  245   &         121   &        3077   &         173   &          0.49   &   12.56   &          0.71   &         765   &         572   &       14704   &   901   &          0.75   &        19.22   &         1.18   \\
 \bfseries 42000 $\xrightarrow[\scriptscriptstyle{9000}]{}$ 51000    &   191   &         178   &        3147   &         186   &          0.93   &     16.48   &          0.97   &         956   &         750   &       17851   &   1087   &          0.78   &        18.67   &         1.14     \\
 \bfseries 51000 $\xrightarrow[\scriptscriptstyle{9000}]{}$ 60000   &  135   &         330   &        3681   &         203   &         2.44   &   27.27   &         1.50   &        1091   &        1080   &       21532   &   1290   &          0.99   &        19.74   &         1.18    \\
\hline
\end{tabular}
\end{table*}

\begin{table*}[!t]
\renewcommand{\arraystretch}{1.3}
\newcommand{\tabincell}[2]{\begin{tabular}{@{}#1@{}}#2\end{tabular}}
\caption{Snapshot Results of Training Time for Presented $4$ BLS Algorithms on Added Inputs and the Corresponding Time Ratios, where the  Newly Added
Inputs are More than the Nodes of the Network.} \label{table_example} \centering
\begin{tabular}{|c||c c c c| c c c|c c c c|c c c|}
\hline
 {\bfseries   \tabincell{c}{Number }}
      &\multicolumn{7}{c|}{{\bfseries   \tabincell{c}{ Additional  Training Time }}}
       &\multicolumn{7}{c|}{{\bfseries   \tabincell{c}{Accumulative Training Time }}}
         \\
   {\bfseries   \tabincell{c}{ of }}
      &\multicolumn{4}{c|}{{\bfseries   \tabincell{c}{ Time in  Seconds }}}
      &\multicolumn{3}{c|}{{\bfseries   \tabincell{c}{  Time Ratios  to Exist}}}
       &\multicolumn{4}{c|}{{\bfseries   \tabincell{c}{ Time in  Seconds }}}
      &\multicolumn{3}{c|}{{\bfseries   \tabincell{c}{ Time Ratios to Exist}}}  \\
  {\bfseries   \tabincell{c}{ Input Patterns  }}    &Exist        &Alg.3       &Alg.4           &Alg.8         &Alg.3       &Alg.4           &Alg.8        &Exist        &Alg.3       &Alg.4           &Alg.8         &Alg.3       &Alg.4           &Alg.8           \\
\hline
  \bfseries  10000  &              11.81    &   12.96    &  103.7    &    6.21    &    1.10   &    8.78    &    0.53    &   11.81    &   12.96    &  103.7    &    6.21    &    1.10   &    8.78    &    0.53\\
 \bfseries  10000 $\xrightarrow[\scriptscriptstyle{10000}]{}$ 20000  &       11.96    &   12.81    &  103.2    &    8.09    &    1.07    &    8.63    &    0.68    &   23.77    &   25.76    &  206.9    &   14.30   &    1.08    &    8.70   &    0.60     \\
 \bfseries 20000 $\xrightarrow[\scriptscriptstyle{10000}]{}$ 30000   &   11.86    &   12.76    &  103.5    &    7.86    &    1.08    &    8.73    &    0.66    &   35.63    &   38.52    &  310.4    &   22.16    &    1.08    &    8.71    &    0.62   \\
 \bfseries 30000 $\xrightarrow[\scriptscriptstyle{10000}]{}$ 40000    &   11.87    &   12.78    &  103.4    &    7.78    &    1.08    &    8.71    &    0.66    &   47.50   &   51.30   &  413.8    &   29.94    &    1.08    &    8.71    &    0.63   \\
 \bfseries 40000 $\xrightarrow[\scriptscriptstyle{10000}]{}$ 50000    &    11.79    &   13.04    &  103.1    &    7.80   &    1.11    &    8.74    &    0.66    &   59.29    &   64.35    &  516.9    &   37.74    &    1.09    &    8.72    &    0.64     \\
 \bfseries 50000 $\xrightarrow[\scriptscriptstyle{10000}]{}$ 60000   &  11.83    &   12.76    &  103.4    &    7.76    &    1.08    &    8.74    &    0.66    &   71.12    &   77.11    &  620.3    &   45.50   &    1.08    &    8.72    &    0.64   \\
\hline
\end{tabular}
\end{table*}

 \begin{table*}[!t]
\renewcommand{\arraystretch}{1.3}
\newcommand{\tabincell}[2]{\begin{tabular}{@{}#1@{}}#2\end{tabular}}
\caption{Snapshot Results of Training Time for Proposed $2$ BLS Algorithms on Added Nodes and the Corresponding Time Ratios} \label{table_example} \centering
\begin{tabular}{|c||c c | c |c c |c |}
\hline
 {\bfseries   \tabincell{c}{Number }}
      &\multicolumn{3}{c|}{{\bfseries   \tabincell{c}{ Additional  Training Time }}}
       &\multicolumn{3}{c|}{{\bfseries   \tabincell{c}{Accumulative Training Time }}}
         \\
   {\bfseries   \tabincell{c}{ of }}
      &\multicolumn{2}{c|}{{\bfseries   \tabincell{c}{ Time in  Seconds }}}
      & {{\bfseries   \tabincell{c}{Speedups of Alg. 8}}}
       &\multicolumn{2}{c|}{{\bfseries   \tabincell{c}{ Time in  Seconds }}}
      & {{\bfseries   \tabincell{c}{Speedups of Alg. 8}}}  \\
  {\bfseries   \tabincell{c}{ Total Nodes  }}       &Alg. 5        &Alg. 8        &  {{\bfseries   \tabincell{c}{over Alg. 5}}}      &Alg. 5        &Alg. 8    & {{\bfseries   \tabincell{c}{over Alg. 5}}}         \\
\hline
  \bfseries 3060  &  28.26  & 6.32   & 4.47  &  28.26  & 6.32   & 4.47    \\
 \bfseries 3060 $\xrightarrow[\scriptscriptstyle{2010}]{}$ 5070  & 207.85  & 33.47   & 6.21  & 236.12   &39.79   & 5.93     \\
 \bfseries 5070 $\xrightarrow[\scriptscriptstyle{2010}]{}$ 7080   & 640.67  & 169.68   & 3.77  & 876.81   &209.47   & 4.18    \\
 \bfseries 7080 $\xrightarrow[\scriptscriptstyle{2010}]{}$ 9090   & 44694  & 1711   & 26.12 & 45571  & 1920  &  23.73   \\
 \bfseries 9090 $\xrightarrow[\scriptscriptstyle{2010}]{}$ 11100    & 90177   &4048    &22.27  &135750  &5968    &22.74     \\
\hline
\end{tabular}
\end{table*}

We simulate the
presented
 low-memory BLS implementations
 on MATLAB software platform under a standard desktop PC (Intel Core i5 Quad-Core with 8 GB DDR4).
 As
  in \cite{BL_trans_paper} and
  \cite{BLSLowMemTNNLS2020dec},
we give the experimental results on the
Modified National Institute of Standards and Technology (MNIST)
dataset~\cite{61_dataSet} with $60000$ training images and $10000$ testing images.
We follow the simulations in the original BLS~\cite{BL_trans_paper}:
for the feature nodes ${{\mathbf{Z}}_{i}}$ in (\ref{Z2PhyXWb985498}),
we fine-tune the random ${{\mathbf{W}}_{{{e}_{i}}}}$ and ${{\mathbf{\beta }}_{{{e}_{i}}}}$
 by the linear inverse problem~\cite{BL_trans_paper},
 and for the
   enhancement nodes ${{\mathbf{H}}_{j}}$ in  (\ref{HjipsenZjWbelta09885}),
   we  choose tansig  for the activation function
    $\xi$.
    The weights
    ${{\mathbf{W}}_{{{h}_{j}}}}$ and
   the biases
    ${{\mathbf{\beta }}_{{{h}_{j}}}}$ ($j=1,2,\cdots, m$)
    in  (\ref{HjipsenZjWbelta09885})
    are drawn from the
standard uniform distributions on the interval $\left[ {\begin{array}{*{20}{c}}
{{\rm{ - }}1}&1
\end{array}} \right]$, and so are the initial
random ${{\mathbf{W}}_{{{e}_{i}}}}$ and ${{\mathbf{\beta }}_{{{e}_{i}}}}$  in (\ref{Z2PhyXWb985498}).

As Table  \Rmnum{5} in \cite{BL_trans_paper}, Table  \Rmnum{1},  Table  \Rmnum{3} and  Table  \Rmnum{4} give the  simulation results for
 the incremental BLS on added inputs. In Table  \Rmnum{1} and  Table  \Rmnum{3},   we set 
 the total node number of the network as  $k=11100$ with 
   $10 \times 10$
feature nodes and $11000$ enhancement nodes,  train the initial network
under the first $l=15000$ training samples,  
and increase $p=9000<k$
training samples 
 in each update, until all the
$60000$ training samples are fed.  
In Table  \Rmnum{4},   we set
 the total node number of the network as  $k=3100$ with
   $10 \times 10$
feature nodes and $3000$ enhancement nodes,  train the initial network
under the first $l=10000$ training samples,
and increase $p=10000>k$
training samples 
 in each update, until all the
$60000$ training samples are fed. 
For each of the above-described update,  the snapshot results  are given in Table  \Rmnum{1}, Table  \Rmnum{3} and  Table  \Rmnum{4}.

 On the other hand,  as Table  \Rmnum{4} in \cite{BL_trans_paper},
 Table  \Rmnum{2} and Table  \Rmnum{5} give the  simulation results for
 the incremental BLS on added nodes.
 We set the initial network as $10 \times 6$
feature nodes and $3000$ enhancement nodes. The feature nodes are dynamically increased
from $60$ to $100$, and the enhancement nodes are dynamically increased
from $3000$ to $11000$. In each update,  $10$ feature nodes are added,
and $2000$ enhancement nodes are added, which include $750$ enhancement nodes  corresponding to the added feature nodes
and
 $1250$ additional enhancement nodes.  
 For each of the above-described update,  the snapshot results  are given in Table  \Rmnum{2} and Table  \Rmnum{5}.

For the presented
 low-memory BLS implementations,
Table \Rmnum{1} and Table \Rmnum{2} show the testing accuracy with the batch size $b=500$.
From Table \Rmnum{1}, it can be seen that
 when the ridge parameter is set to  $\lambda=1/128$ as in \cite{BLSLowMemTNNLS2020dec},
 the proposed \textbf{Algorithm 3},  \textbf{Algorithm 4} and the part for added inputs in \textbf{Algorithm 8}   usually achieve
 better testing accuracies than the existing low-memory  BLS implementation on added inputs  proposed
 in \cite{BLSLowMemTNNLS2020dec}. More importantly,  Table \Rmnum{1} shows that
 when the ridge parameter
  is very small (i.e., $\lambda = 10^{-8}$) as in the original BLS~\cite{BL_trans_paper},
  the existing low-memory  BLS implementation on added inputs  proposed
 in \cite{BLSLowMemTNNLS2020dec} cannot work in any update,
  the proposed recursive implementation (i.e., \textbf{Algorithm 3})  cannot work in the last update,
  and the proposed square-root implementations  (i.e., \textbf{Algorithm 4} and the part for added inputs in \textbf{Algorithm 8}) can work in all updates.
  This can be explained by the fact that both
  the proposed recursive BLS implementation and the BLS implementation in \cite{BLSLowMemTNNLS2020dec}
  include the recursive updates of
    the inverse matrix, which may introduce numerical instabilities
   after a large number of iterations in the processor units with limited precision~\cite{TransSP2003Blast}.
   On the other hand,   Table \Rmnum{2} shows that both the proposed low-memory square-root  BLS implementations on added nodes
   can work for a small ridge parameter  $\lambda = 10^{-7}$, while the part for added nodes in \textbf{Algorithm 8}
   can work for the very small ridge parameter  $\lambda = 10^{-8}$ utilized in the original BLS~\cite{BL_trans_paper}.

   Table \Rmnum{3},  Table \Rmnum{4} and Table \Rmnum{5} include the  training times in seconds with the batch size $b=50$.
    Table \Rmnum{3} and Table \Rmnum{4} show that with respect to the existing low-memory  BLS implementation on added inputs  proposed
 in \cite{BLSLowMemTNNLS2020dec},    \textbf{Algorithm 3} and the part for added nodes in \textbf{Algorithm 8} require nearly
 the same training time, while \textbf{Algorithm 4} requires much more training time.  On the other hand,
  Table \Rmnum{5} shows that the part for added nodes in \textbf{Algorithm 8} speeds up
  \textbf{Algorithm 5} by a factor in the range from $3.77$ to $23.73$.

 From Tables \Rmnum{1},  \Rmnum{2},  \Rmnum{3},  \Rmnum{4} and \Rmnum{5},  it can be concluded that the
 proposed  \textbf{Algorithm 8} is a good low-memory  implementation of the original BLS on new added
 nodes and inputs~\cite{BL_trans_paper}, which can work when the ridge parameter
  is very small (i.e., $\lambda = 10^{-8}$) as in the original BLS.
  The part for added inputs in \textbf{Algorithm 8}
  spends nearly
 the same training time to
   achieve
 better testing accuracies
   with respect to  the existing low-memory  BLS implementation on added inputs~\cite{BLSLowMemTNNLS2020dec},
 is numerically more stable  than \textbf{Algorithm 3}  (i.e.,   the proposed recursive implementation),
 and is much faster than  \textbf{Algorithm 4}. On the other hand,
  the part for added nodes in \textbf{Algorithm 8} is obviously faster than \textbf{Algorithm 5}.

 \section{Conclusion}

In this paper, firstly we propose the low-memory  implementations
for
the recursive and square-root BLS algorithms on new added inputs in \cite{mybrief2onBLInputs2019}
and the square-root BLS algorithm on new added nodes in \cite{mybrief1onBLNodes2019}, which  simply process a batch of $b$ inputs or nodes in each recursion  by
 a $b \times b$ matrix inversion or inverse Cholesky factorization.
 Since the recursive BLS implementation includes
 the recursive updates of
       the inverse matrix that
may introduce numerical instabilities
   after a large number of iterations~\cite{TransSP2003Blast},
   and needs the extra computational load to decompose the
 inverse matrix  into the Cholesky factor  when cooperating with the proposed low-memory implementation of
 the square-root BLS algorithm on added nodes in \cite{mybrief1onBLNodes2019},
 we only improve the  low-memory implementations of the square-root BLS algorithms on added inputs and nodes,
 to propose the full low-memory implementation of the square-root BLS algorithm.

  When the ridge parameter is set to  $\lambda=1/128$ as in \cite{BLSLowMemTNNLS2020dec},
 the proposed low-memory  implementations for
the recursive and square-root BLS algorithms on added inputs in \cite{mybrief2onBLInputs2019} and the part for added inputs
of the proposed full low-memory implementation of the square-root BLS algorithm
 usually achieve
 better testing accuracies than the existing low-memory  BLS implementation on added inputs  proposed
 in \cite{BLSLowMemTNNLS2020dec}.
 More importantly,
 when the ridge parameter
  is very small (i.e., $\lambda = 10^{-8}$) as in the original BLS~\cite{BL_trans_paper},
  the existing low-memory  BLS implementation on added inputs  proposed
 in \cite{BLSLowMemTNNLS2020dec} cannot work in any update,
 the proposed low-memory  implementation for
the recursive BLS algorithm on added inputs in \cite{mybrief2onBLInputs2019}
  cannot work in the last update,
  while  the proposed low-memory  implementation for
the square-root BLS algorithm on added inputs in \cite{mybrief2onBLInputs2019}
and the proposed full low-memory implementation of the square-root BLS algorithm on added inputs and nodes
   can work in all updates.

    With respect to the existing low-memory  BLS implementation on added inputs  proposed
 in \cite{BLSLowMemTNNLS2020dec},
 the proposed low-memory  implementation for
the recursive BLS algorithm on added inputs in \cite{mybrief2onBLInputs2019}
  and  the part for added inputs
of the proposed full low-memory implementation of the square-root BLS algorithm
   require nearly
 the same training time, while the proposed low-memory  implementation for
the square-root BLS algorithm on added inputs in \cite{mybrief2onBLInputs2019} requires much more training time. On the other hand,
   the part for added nodes of  the proposed full low-memory implementation of the square-root BLS algorithm speeds up
 the  proposed low-memory  implementation
for the square-root BLS algorithm on new added nodes in \cite{mybrief1onBLNodes2019}
   by a factor in the range from $3.77$ to $23.73$.

The  proposed  full low-memory implementation of the square-root BLS algorithm on added inputs and nodes
can work when the ridge parameter
  is very small (i.e., $\lambda = 10^{-8}$) as in the original BLS.
The part for added inputs of  the  proposed  full low-memory implementation of the square-root BLS algorithm
  takes nearly
 the same training time to
   achieve
 better testing accuracies
   with respect to  the existing low-memory  BLS implementation on added inputs~\cite{BLSLowMemTNNLS2020dec},
    is numerically more stable  than  the proposed low-memory  implementation for
the recursive BLS algorithm on added inputs in \cite{mybrief2onBLInputs2019},
 and is much faster than  the proposed low-memory  implementation for
the square-root BLS algorithm on added inputs in \cite{mybrief2onBLInputs2019}.
On the other hand, the part for added nodes of  the  proposed  full low-memory implementation of the square-root BLS algorithm  is obviously faster than
the  proposed low-memory  implementation
for the square-root BLS algorithm on added nodes in \cite{mybrief1onBLNodes2019}.
Accordingly,
it can be concluded that
  the  proposed  full low-memory implementation of the square-root BLS algorithm on added inputs and nodes
  is
  a good low-memory  implementation of the original BLS on added
 nodes and inputs~\cite{BL_trans_paper}.

\appendices

\ifCLASSOPTIONcaptionsoff
  \newpage
\fi

%

%
%
%




\end{document}